\documentclass[12pt,a4paper,twoside]{article}
\makeatletter\if@twocolumn\PassOptionsToPackage{switch}{lineno}\else\fi\makeatother

\usepackage{amsfonts,amssymb,amsbsy,latexsym,amsmath,tabulary,graphicx,times,xcolor}
\usepackage[T1]{fontenc}
\usepackage{url,multirow,morefloats,floatflt,cancel,tfrupee,multicol}
\makeatletter
 
\AtBeginDocument{\@ifpackageloaded{textcomp}{}{\usepackage{textcomp}}}
\makeatother
\usepackage{colortbl}
\usepackage{xcolor}
\usepackage{pifont}
\usepackage[nointegrals]{wasysym}
\usepackage[superscript,biblabel]{cite}
\makeatletter

\usepackage{graphicx}
\usepackage{pgfplots}
\usepackage{tikz}
\usepgfplotslibrary{external} 
\usetikzlibrary{external,calc,arrows.meta,patterns,positioning} 
\DeclareMathOperator{\arctantwo}{arctan2}

\@ifundefined{etal}{}{}

\AtBeginDocument{
\expandafter\ifx\csname eqalign\endcsname\relax
\def\eqalign#1{\null\vcenter{\def\\{\cr}\openup\jot\m@th
  \ialign{\strut$\displaystyle{##}$\hfil&$\displaystyle{{}##}$\hfil
      \crcr#1\crcr}}\,}
\fi
}

\@ifundefined{tsGraphicsScaleX}{\gdef\tsGraphicsScaleX{1}}{}
\@ifundefined{tsGraphicsScaleY}{\gdef\tsGraphicsScaleY{.9}}{}
\def\checkGraphicsWidth{\ifdim\Gin@nat@width>\linewidth
	\tsGraphicsScaleX\linewidth\else\Gin@nat@width\fi}

\def\checkGraphicsHeight{\ifdim\Gin@nat@height>.9\textheight
	\tsGraphicsScaleY\textheight\else\Gin@nat@height\fi}

\def\fixFloatSize#1{}
\let\ts@includegraphics\includegraphics

\def\inlinegraphic[#1]#2{{\edef\@tempa{#1}\edef\baseline@shift{\ifx\@tempa\@empty0\else#1\fi}\edef\tempZ{\the\numexpr(\numexpr(\baseline@shift*\f@size/100))}\protect\raisebox{\tempZ pt}{\ts@includegraphics{#2}}}}

\AtBeginDocument{\def\includegraphics{\@ifnextchar[{\ts@includegraphics}{\ts@includegraphics[width=\checkGraphicsWidth,height=\checkGraphicsHeight,keepaspectratio]}}}

\DeclareMathAlphabet{\mathpzc}{OT1}{pzc}{m}{it}

\def\URL#1#2{\@ifundefined{href}{#2}{\href{#1}{#2}}}

\edef\fntEncoding{\f@encoding}

\makeatother

\newif\ifmultipleabstract\multipleabstractfalse%
\usepackage[hmargin=1.8cm,vmargin=2.2cm]{geometry}
\makeatletter
\def\author#1{\gdef\@author{\hskip-\dimexpr(\tabcolsep)\hskip1pt\parbox{\dimexpr\textwidth-1pt}{\centering #1}}}

\let\@articletype\@empty \def\articletype#1{\gdef\@articletype{{\fontsize{14}{16}\selectfont #1}}}

\usepackage{fancyhdr}

\fancypagestyle{headings}{\fancyhf{}\fancyhead[RE]{\MakeTextUppercase{\textbf{\RunningAuthor}}}\fancyhead[LE]{\textbf{\thepage}}\fancyhead[LO]{\MakeTextUppercase{\textbf{\RunningHead}}}\fancyhead[RO]{\textbf{\thepage}}}
\fancypagestyle{plain}{\fancyhf{}\fancyfoot[C]{\textbf{\thepage}}}

\AtBeginDocument{\usepackage{lastpage}}
\def\title#1{%
  \gdef\@title{%
    \ifx\@articletype\@empty\else\@articletype~\\\fi%
     #1}%
}
\usepackage[font=footnotesize]{caption}
\usepackage[font=footnotesize]{subcaption}

\usepackage{abstract}
\def\abstractname{\textbf{Abstract}}
\renewenvironment{onecolabstract}
{\vspace*{-.4pc}\trivlist\item[]\leftskip1pt\noindent\selectfont\hfill\abstractname\hfill\mbox{\null}\par\ignorespaces}{\endtrivlist}

\def\NormalBaseline{\def\baselinestretch{1.1}}

\usepackage{textcase}
\usepackage[explicit]{titlesec}
\titleformat{\section}[block]{\NormalBaseline\boldmath\bfseries}
{\thesection.}
{6pt}
{#1}
[]
\titleformat{\subsection}[hang]{\NormalBaseline\filright\itshape}
{\thesubsection.}
{6pt}
{#1}
[]
\titleformat{\subsubsection}[runin]{\NormalBaseline\filright\itshape}
{\hspace{16pt}\thesubsubsection}
{6pt}
{#1\newline}
[]
\titleformat{\paragraph}[runin]{\NormalBaseline}
{\theparagraph}
{6pt}
{#1}
[]
\titleformat{\subparagraph}[runin]{\NormalBaseline}
{\thesubparagraph}
{6pt}
{#1}
[]

\titlespacing{\section}{0pt}{1.5\baselineskip}{.2\baselineskip}  
\titlespacing{\subsection}{0pt}{1.5\baselineskip}{.2\baselineskip}  
\titlespacing{\subsubsection}{0pt}{1.5\baselineskip}{.2\baselineskip}  
\titlespacing{\paragraph}{0pt}{.5\baselineskip}{10pt}  
\titlespacing{\subparagraph}{0pt}{.5\baselineskip}{10pt}

\usepackage{caption}
\DeclareCaptionLabelFormat{figlabel}{Figure #2}
\date{}
\makeatother

\usepackage{float}

\begin{document}

\title{Sim-to-real for high-resolution optical tactile sensing:\\ From images to 3D contact force distributions}
\def\RunningHead{
Sim-to-real for high-resolution optical tactile sensing
}
\def\RunningAuthor{Sferrazza and D'Andrea}
\author{Carmelo Sferrazza and Raffaello D'Andrea
\thanks{Institute for Dynamic Systems and Control, ETH Zurich, 8092 Zurich, Switzerland. \newline\hspace*{6mm} Correspondence to:
        {\tt\small csferrazza@ethz.ch}}%
}

\maketitle

\widowpenalty10000
\clubpenalty10000

%
\begin{figure}[H]
	\centering
	\vspace{-0.cm}
	\subcaptionbox{Simulated indentation}{
		\setlength{\fboxsep}{0pt}
		\fbox{\includegraphics[height=0.2\columnwidth]{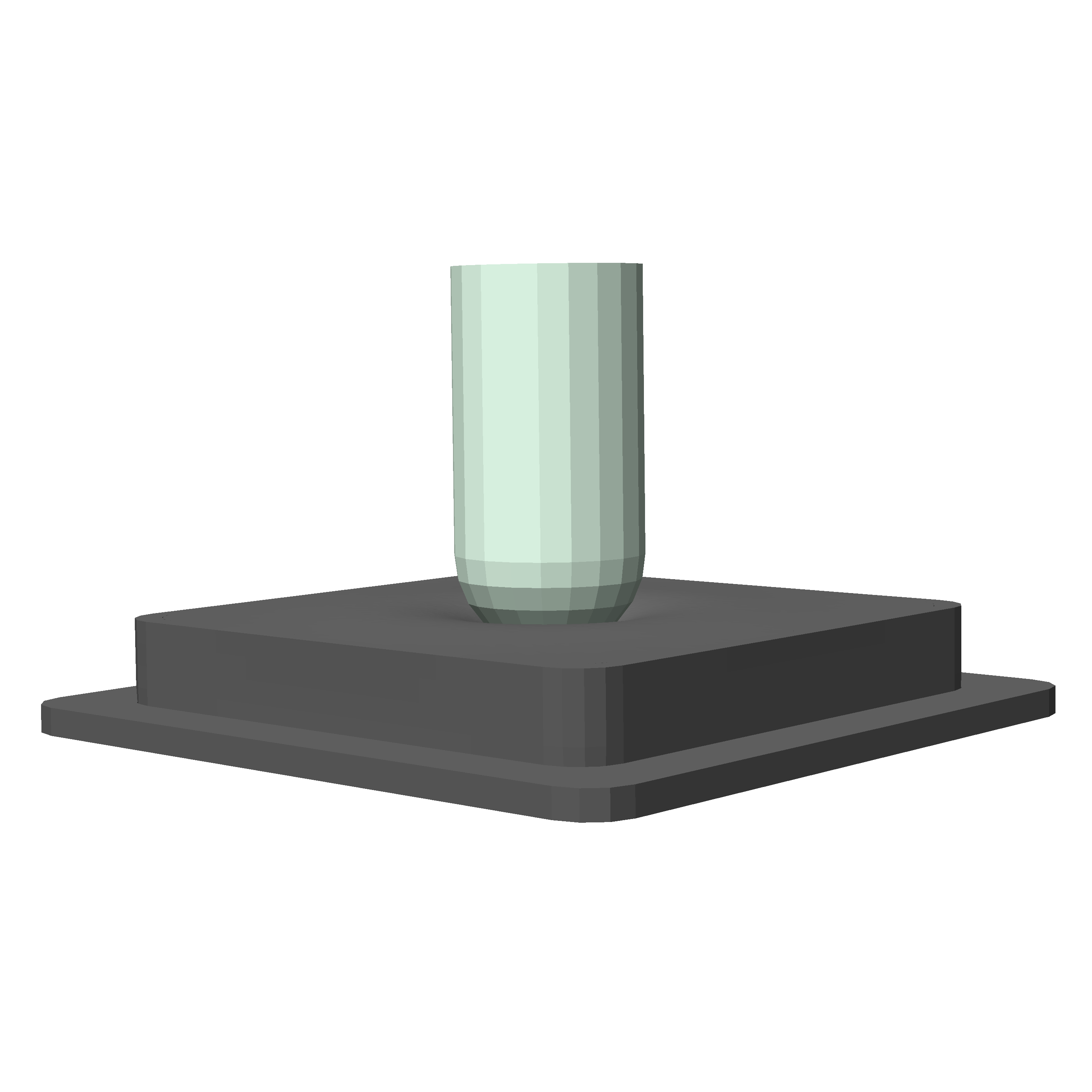}}
	} 
	\subcaptionbox{Real-world indentation}{
		\setlength{\fboxsep}{0pt}
		\fbox{\includegraphics[height=0.2\columnwidth]{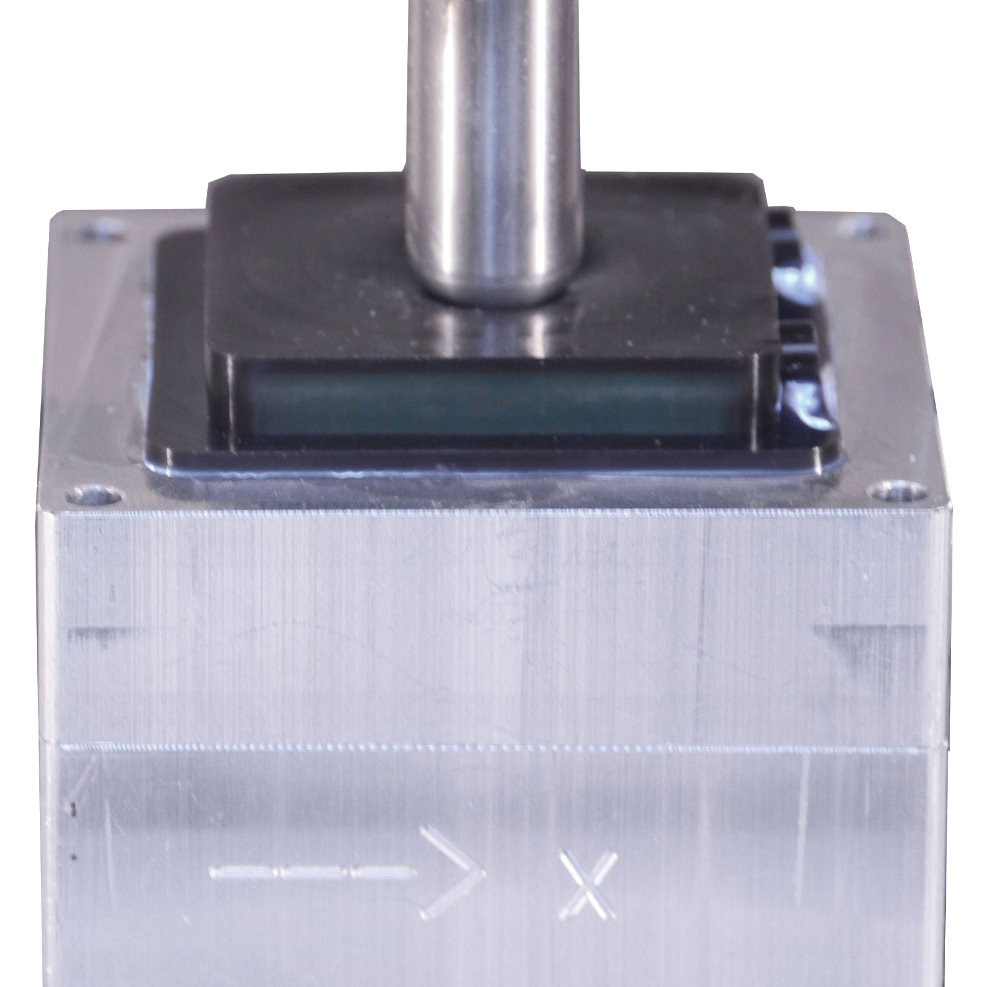}}
	}
	\vskip \baselineskip
	\subcaptionbox{Simulated image}{
		\setlength{\fboxsep}{0pt}
		\fbox{\includegraphics[height=0.2\columnwidth]{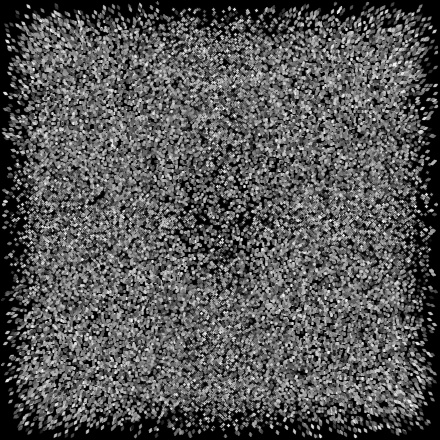}}
	} 
	\subcaptionbox{Real-world image}{
		\setlength{\fboxsep}{0pt}
		\fbox{\includegraphics[height=0.2\columnwidth]{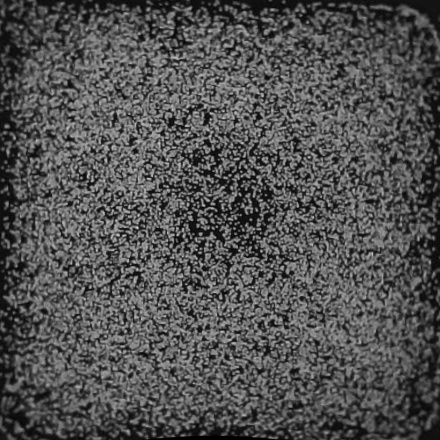}}
	}
	\caption{This work builds upon the generation of training images in simulation for a data-driven, vision-based tactile sensor based on the tracking of a spread of particles.}
	\label{fig:teaser}
\end{figure}
{\begin{onecolabstract}
The images captured by vision-based tactile sensors carry information about high-resolution tactile fields, such as the distribution of the contact forces applied to their soft sensing surface. However, extracting the information encoded in the images is challenging and often addressed with learning-based approaches, which generally require a large amount of training data. This article proposes a strategy to generate tactile images in simulation for a vision-based tactile sensor based on an internal camera that tracks the motion of spherical particles within a soft material. The deformation of the material is simulated in a finite element environment under a diverse set of contact conditions, and spherical particles are projected to a simulated image. Features extracted from the images are mapped to the 3D contact force distribution, with the ground truth also obtained via finite-element simulations, with an artificial neural network that is therefore entirely trained on synthetic data avoiding the need for real-world data collection. The resulting model exhibits high accuracy when evaluated on real-world tactile images, is transferable across multiple tactile sensors without further training, and is suitable for efficient real-time inference.

\def\keywordstitle{Keywords}
\smallskip\noindent\textbf{Keywords: }{\normalfont
tactile sensing, sim-to-real, machine learning, computer vision
}
\end{onecolabstract}}
 
\begin{multicols}{2}

\section{Introduction} \label{sec:introduction}
Research on vision-based (or optical) tactile sensors aims to provide robots with high-resolution information about contact with external objects. However, while the images stemming from the various optical tactile sensing principles are intuitive and to some extent interpretable by human observations, the extraction of accurate physical quantities is challenging. In this regard, the complexity of mapping the information extracted from the images to the corresponding contact conditions mainly results from the fact that accurate modeling techniques for soft materials are generally not suitable for real-time applications. Additionally, previous research has predominantly focused on the estimation of low-dimensional quantities (e.g., total contact force, center of contact), which may be sufficient for a limited range of tasks, but not for generic applications, as is the case for tasks that involve arbitrary points of contact.

The work discussed in this article targets both these topics, proposing a data-driven approach to reconstruct the three-dimensional distribution of the contact forces applied to the soft surface of a vision-based tactile sensor. The sensing strategy was presented in the authors' previous work\cite{sferrazza_sensors} and is based on the tracking of particles randomly spread within a soft gel. The use of data bypasses the need for modeling techniques with real-time guarantees, but as opposed to classical data-driven strategies, here the data necessary for training the learning architecture at the core of the method are entirely generated in simulation. Furthermore, the estimation of the contact force distribution directly yields both the total contact force (i.e., the component-wise integral of the force distribution) and the contact locations (i.e., the surface patches where the contact pressure is nonzero), and is additionally suitable to represent generic contact conditions with arbitrary points of contact, therefore providing high versatility across several tasks.

The main contributions of this work are the following:
\begin{enumerate}
	\item[$\bullet$] It details a method to simulate the images captured by a vision-based tactile sensor\cite{sferrazza_sensors}, starting from simulations based on the finite element method\cite{fem_book} (FEM). 
	\item[$\bullet$] It outlines two strategies to generate simulated datasets comprising tactile image features and labels. These strategies differ from the one presented in the authors' previous work,\cite{sim2real_iros} as they relax a small deformation assumption and simplify the transfer from simulation to reality. The datasets collected for this work comprise a variety of contact conditions, producing high shear and pressure forces with indenters of different shapes and sizes. 
	\item[$\bullet$] It describes a tailored learning architecture, based on u-net\cite{unet_paper}, which can be trained entirely with simulated data obtained offline via high-fidelity FEM-based simulations. When evaluated on real-world tactile sensors, the architecture yields high accuracy in the reconstruction of the force distribution, achieving real-time inference up to a speed of 120 Hz.  
\end{enumerate}

\subsection{Related work}
In recent years, a number of tactile sensing principles\cite{survey_humanoids} have been developed to address the needs of the robotics community. Among these, vision-based tactile sensors\cite{survey_kazu} employ standard cameras\cite{tactip_family} or optical devices\cite{davis_2020,survey_gelsight} to infer the deformation of a soft membrane and obtain information about the contact with external objects that causes the deformation. This category of tactile sensors generally benefits from high resolution and ease of wiring, and its straightforward manufacture enables fast prototyping for robotic systems. Although the bulkiness of their sensing unit is the main limitation of such approaches, recent works have proposed compact solutions that exploit embedded cameras\cite{camill_robosoft,digit_paper,omnitact_paper,round_gelsight,geltip_paper} or mirrors.\cite{gelslim_paper}

The sensory feedback provided by tactile sensors typically requires further processing, as it does not directly translate to the physical quantities of interest for robotic tasks. In this regard, model-based methods\cite{gelslim_fem, softbubble_fem} often rely on strong modeling assumptions (e.g., linear elasticity of the materials) to solve the processing task in an approximate fashion, while data-driven methods\cite{sferrazza_fem,gelsight_sensors,piacenza_tom} aim to compute offline a mapping from raw data to the quantities of interest, in order to preserve accuracy while ensuring real-time inference.

While most of the literature has primarily focused on the estimation of low-dimensional physical quantities (e.g., total forces), recently several works have shifted the focus towards the estimation of distributed quantities, which aim to provide high-resolution tactile fields for a wide range of tasks. In the context of vision-based sensors, the estimation of the contact patches\cite{softbubble_fem} has been proposed, and the reconstruction of the contact force distribution has been discussed, both in a model-based\cite{gelslim_fem} and a data-driven\cite{sferrazza_fem} fashion. Additionally, various approaches have been proposed outside the vision-based domain, with regard to the estimation of the deformation field\cite{yash_fem} and the pressure distribution.\cite{hyosang_fem}

As a result of the possibility of collecting and generating accurate data offline, data-driven approaches generally exhibit smaller estimation errors than model-based methods.\cite{davis_2020} However, their bottleneck often lies in the fact that they require large amounts of training data and they do not often generalize well when employed in unseen contact conditions. In order to address the issue of data efficiency, a number of works have focused on generating training data in simulation to extract a model that retains its accuracy when employed in the real world. Examples of such sim-to-real (or sim2real) transfers can be found in the literature for edge prediction\cite{lepora_sim2real} and the estimation of the contact pressure\cite{hyosang_fem} and the deformation field.\cite{vitac_sim2real,elastic_particles} In previous work, a sim-to-real approach was presented to estimate the 3D force distribution\cite{sim2real_iros} for a limited range of scenarios.

This article presents two different methods to generate a dataset to train a data-driven approach entirely via FEM simulations, with the aim to reconstruct the three-dimensional contact force distribution applied to a vision-based tactile sensor. Image features were extracted from the tactile images generated in simulation, and mapped to three matrices representing the components of the force vectors applied over the soft sensing surface. The mapping was obtained via a tailored neural network architecture, which is able to capture various contact conditions as high shear and pressure forces, as well as indentations with flat or round objects. Additionally, high accuracy was retained on real-world data and the real-time speed could be more than doubled compared to previous work.\cite{sim2real_iros}

\subsection{Outline}
The sensing strategy and the hardware are described in Section \ref{sec:hardware}, while the method to generate tactile images and extract the related features is presented in Section \ref{sec:dataset_generation}. Starting from the generated dataset, Sections \ref{sec:learning} and \ref{sec:results} describe the learning pipeline and the evaluation on simulated and real-world data, respectively. Final remarks and an outlook are included in Section \ref{sec:conclusion}.

\section{Materials and Methods}
\subsection{Hardware} \label{sec:hardware}
The tactile sensor employed in this work is based on a camera that tracks particles randomly distributed within a deformable material. The fabrication follows previous work,\cite{sferrazza_sensors} and is detailed in Section 1 of the supplementary material. The sensing surface amounts to a rectangular prism of 32$\times$32$\times$6 mm. The soft materials have been characterized previously\cite{sferrazza_fem} as hyperelastic materials following uniaxial, pure shear and equibiaxial tension tests. The resulting second-order Ogden models\cite{ogden_model} were employed for the FEM simulations discussed in the following sections.
\subsection{Dataset generation} \label{sec:dataset_generation}
Supervised learning is a natural data-driven way of processing sensory feedback and mapping raw data to the quantities of interest. In the context of vision-based tactile sensing, formulating the task in a supervised learning manner involves two crucial preliminary steps: I) the choice of appropriate features to condense the information contained in the images; II) the formalization of finite-dimensional labels representing the quantities of interest. Additionally, the availability of data necessary to train suitable learning architectures needs to be considered when addressing the formulation of the problem. In this work, training data were generated entirely in a finite element simulation environment with the objective of avoiding real-world data collection and maximizing the variability of the contact conditions without the need for complex hardware setups. A further advantage of collecting contact data in simulation is the possibility of extracting high-resolution tactile fields,\cite{sferrazza_fem} which are otherwise not possible to measure with the commercially available commodity sensors. This work aimed to estimate the three-dimensional force distribution, which is a condensed representation of several contact quantities. In fact, the force distribution encodes both the contact locations, which can be obtained by thresholding the normal component, and the total contact forces, which can be obtained by integrating the distribution over the sensing surface. As opposed to the deformation field, the contact patches are exactly encoded in the force distribution, while the deformation field can, for example, show deformation also where no contact is applied, as a result of the elasticity of the soft material. Additionally, from the force distribution it is possible to compute the torques acting on the contact object, and all these properties remain valid for contact with multiple or arbitrary objects.

\begin{figure}[H]
	\centering
	\includegraphics[width=\columnwidth]{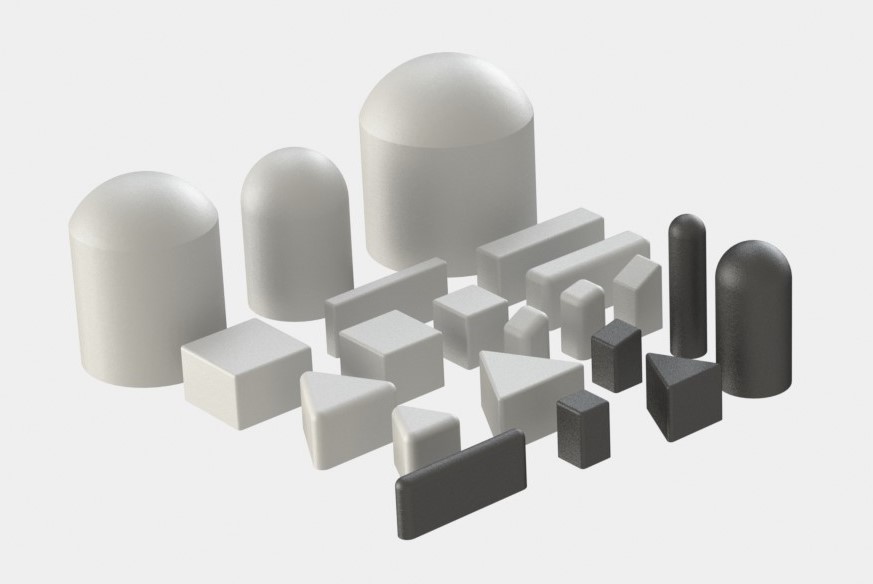}
	\caption{The figure shows the indenters used to collect the training data in the FEM simulations. Real-world realizations of the black indenters were used to collect the test data in reality. Note that the indentation surfaces correspond to the top surfaces in the figure. The sharp corners of the indenters were smoothed out to avoid a known singularity in the flat-punch indentation experiment.\cite{flat_punch}}
	\label{fig:indenters}
\end{figure}
%

An FEM simulation environment was created in Abaqus/Standard\cite{abaqus_manual}, details of this are provided in Section 2 of the supplementary material and in a previous work\cite{sferrazza_fem}. Two training datasets were built by performing indentations in such an FEM environment with the 21 different indenters shown in Fig.~\ref{fig:indenters}. The indentation trajectories were performed by either moving the indenter vertically and then purely horizontally, or by prescribing indenter motions from different angles followed by random perturbations in the vicinity of the first indentation. A total of 3300 indentation trajectories (each comprising 50 indentation steps) were executed in simulation, with total forces up to 16 N in the vertical direction and up to 5 N in each of the horizontal directions. For each step of these trajectories, the contact force distribution and the displacement field were extracted at the nodes of a mesh refined around the contact between the indenter and the soft material. These quantities were further processed to compose two sets of features and labels, as described in \mbox{Section \ref{subsec:training_features}} for the displacement field and \mbox{Section \ref{subsec:training_labels}} for the force distribution. Since the training dataset was entirely generated in simulation, two test datasets were collected in reality as described in \mbox{Section \ref{subsec:test_set}} to verify the sim-to-real transfer and the real-world performance.

\subsubsection{Training features}
\label{subsec:training_features}
In this article, two different methods to extract image features are compared. The resulting types of features are denoted in the following as optical flow features and raw features, respectively. The starting points of both methods are the images captured by the internal camera, and for training purposes, these images were entirely generated in simulation. The soft materials were modeled in the FEM simulations as described in Section \ref{sec:hardware} of this manuscript and in Section 2 of the supplementary material. Highly accurate models were obtained for the same materials via state-of-the-art characterization experiments in previous work,\cite{sferrazza_fem} where these models were also validated against a force-torque sensor. The Ogden model parameters used there were also employed in this work. A static friction coefficient of 0.9 was used, as it proved accurate for the indenters employed (see the experiments performed in Section 2 of the supplementary material).

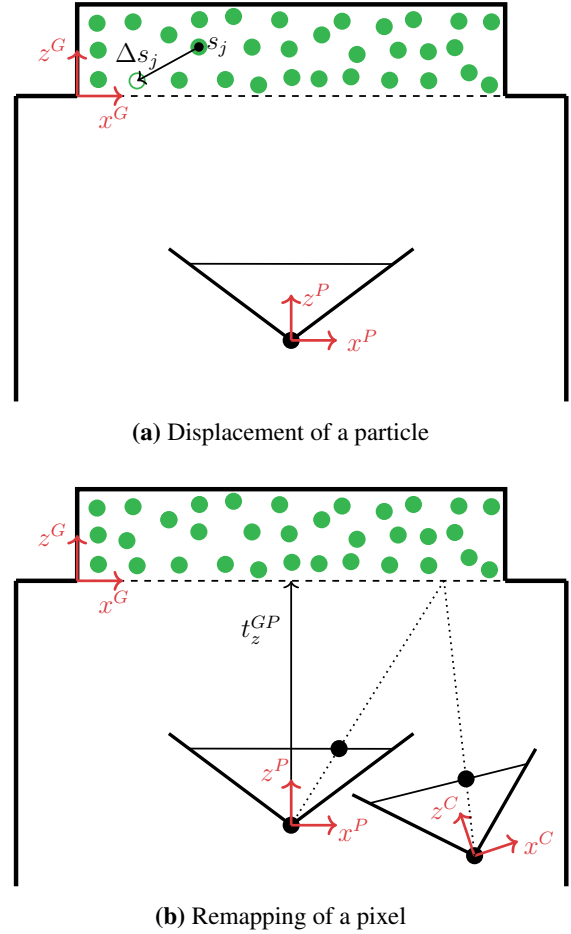
\begin{figure}[H]
\centering
\subcaptionbox{Displacement of a particle}{
\scalebox{0.9}{\def\r{8}
\definecolor{green}{RGB}{54,181,80}
\definecolor{red}{RGB}{220,57,61}
\begin{tikzpicture}[scale=0.15,thick,x=1mm,y=1mm]
  \filldraw[green] (278, 496) circle (\r);
  \filldraw[green] (276.808, 524.2324) circle (\r);
  \filldraw[green] (278.2292, 465.2516) circle (\r);
  \filldraw[green] (362.7921, 464.541) circle (\r);
  \draw[green] (318.7341, 463.8304) circle (\r);
  \filldraw[green] (383.3998, 499.361) circle (\r);
  \filldraw (383.3998, 499.361) circle (4) node[right,xshift=-1] {$s_j$};
  \filldraw[green] (439.0365, 499.2034) circle (\r);
  \filldraw[green] (445.9337, 459.5667) circle (\r);
  \filldraw[green] (480.7537, 467.3834) circle (\r);
  \filldraw[green] (509.1782, 465.9622) circle (\r);
  \filldraw[green] (542.5769, 468.0941) circle (\r);
  \filldraw[green] (477.9112, 497.2292) circle (\r);
  \filldraw[green] (533.339, 526.3643) circle (\r);
  \filldraw[green] (548.9724, 491.5443) circle (\r);
  \filldraw[green] (577.3969, 520.6794) circle (\r);
  \filldraw[green] (583.0818, 464.541) circle (\r);
  \filldraw[green] (623.5867, 494.3867) circle (\r);
  \filldraw[green] (651.3006, 498.6504) circle (\r);
  \filldraw[green] (655.5643, 526.3643) circle (\r);
  \filldraw[green] (689.6737, 525.6536) circle (\r);
  \filldraw[green] (666.2235, 472.3577) circle (\r);
  \filldraw[green] (687.5418, 459.5667) circle (\r);
  \filldraw[green] (419.641, 531.3385) circle (\r);
  \filldraw[green] (467.9627, 527.7855) circle (\r);
  \filldraw[green] (411.1137, 465.2516) circle (\r);
  \filldraw[green] (314.4704, 525.6536) circle (\r);
  \filldraw[green] (384, 528) circle (\r);
  \filldraw[green] (512, 512) circle (\r);
  \filldraw[green] (592, 496) circle (\r);
  \filldraw[green] (608, 528) circle (\r);
  \filldraw[green] (624, 464) circle (\r);
  \filldraw[green] (352, 512) circle (\r);
  
  \filldraw (480, 192) circle (\r);

  \draw [line width = 2.0](192, 128) -- (192, 448);
  \draw [line width = 2.0](192, 448) -- (256, 448);
  \draw [line width = 2.0](256, 448) -- (256, 544) 
  -- (704, 544) -- (704, 448);
  \draw [line width = 2.0](704, 448) -- (768, 448);
  \draw [line width = 2.0](768, 448) -- (768, 128);
  \draw [dashed](256, 448) -- (704, 448);
  \draw [line width = 1.5](480, 192) -- (352, 288);
  \draw [line width = 1.5](480, 192) -- (608, 288);
  \draw (372.9707, 272.2719) -- (586.6296, 271.9722);  
  \draw[red, ->, line width = 1.2] (256, 448) -- node [left,pos=1] {$z^G$} (256, 496);
  \draw[red, ->, line width = 1.2] (256, 448) -- node [below,pos=0.8]{$x^G$} (304, 448);
  \draw[red, ->, line width = 1.2] (480, 192) -- node [right,pos=1]{$z^P$} (480, 240);
  \draw[red, ->, line width = 1.2] (480, 192) -- node [right,pos=1]{$x^P$} (528, 192);
  
  \draw[->] (383.3998, 499.361) -- node [above, pos=1, xshift=0,yshift=0.5]{$\Delta s_j$} (318.7341, 463.8304);

\end{tikzpicture}}
}
\vskip \baselineskip
\subcaptionbox{Remapping of a pixel}{
\scalebox{0.9}{\def\r{8}
\definecolor{green}{RGB}{54,181,80}
\definecolor{red}{RGB}{220,57,61}
\begin{tikzpicture}[scale=0.15,thick,x=1mm,y=1mm]

	\filldraw[green] (278, 496) circle (\r);
	\filldraw[green] (276.808, 524.2324) circle (\r);
	\filldraw[green] (278.2292, 465.2516) circle (\r);
	\filldraw[green] (362.7921, 464.541) circle (\r);
	\filldraw[green] (318.7341, 463.8304) circle (\r);
	\filldraw[green] (383.3998, 499.361) circle (\r);
	\filldraw[green] (439.0365, 499.2034) circle (\r);
	\filldraw[green] (445.9337, 459.5667) circle (\r);
	\filldraw[green] (480.7537, 467.3834) circle (\r);
	\filldraw[green] (509.1782, 465.9622) circle (\r);
	\filldraw[green] (542.5769, 468.0941) circle (\r);
	\filldraw[green] (477.9112, 497.2292) circle (\r);
	\filldraw[green] (533.339, 526.3643) circle (\r);
	\filldraw[green] (548.9724, 491.5443) circle (\r);
	\filldraw[green] (577.3969, 520.6794) circle (\r);
	\filldraw[green] (583.0818, 464.541) circle (\r);
	\filldraw[green] (623.5867, 494.3867) circle (\r);
	\filldraw[green] (651.3006, 498.6504) circle (\r);
	\filldraw[green] (655.5643, 526.3643) circle (\r);
	\filldraw[green] (689.6737, 525.6536) circle (\r);
	\filldraw[green] (666.2235, 472.3577) circle (\r);
	\filldraw[green] (687.5418, 459.5667) circle (\r);
	\filldraw[green] (419.641, 531.3385) circle (\r);
	\filldraw[green] (467.9627, 527.7855) circle (\r);
	\filldraw[green] (411.1137, 465.2516) circle (\r);
	\filldraw[green] (314.4704, 525.6536) circle (\r);
	\filldraw[green] (308.0749, 490.123) circle (\r);
	\filldraw[green] (384, 528) circle (\r);
	\filldraw[green] (512, 512) circle (\r);
	\filldraw[green] (592, 496) circle (\r);
	\filldraw[green] (608, 528) circle (\r);
	\filldraw[green] (624, 464) circle (\r);
	\filldraw[green] (352, 512) circle (\r);
	\filldraw[green] (352, 512) circle (\r);

    \filldraw (480, 192) circle (\r);
    \filldraw (672, 160) circle (\r);
    \filldraw (663.055, 240.5046) circle (\r);	
    \filldraw (530.1421, 272.2273) circle (\r);
	
	\draw[line width = 2] (192, 128) -- (192, 448);
	\draw[line width = 2] (192, 448) -- (256, 448);
	\draw[line width = 2] (256, 448) -- (256, 544) 
	-- (704, 544) -- (704, 448);
	\draw[line width = 2] (704, 448) -- (768, 448);
	\draw[line width = 2] (768, 448) -- (768, 128);
	\draw[dashed] (256, 448) -- (704, 448);
	\draw[line width = 1.5] (480, 192) -- (352, 288);
	\draw[line width = 1.5] (480, 192) -- (608, 288);
	\draw (372.9707, 272.2719) -- (586.6296, 271.9722)
	-- (586.857, 272.1427);


	\draw[line width = 1.5] (672, 160) -- (736, 272);
	\draw[line width = 1.5] (672, 160) -- (544, 224);
	\draw (726.9943, 256.24) -- (561.9927, 215.0036);
	\draw[shift={(671.891, 159.941)}, rotate=18.4441, red, ->,line width = 1.2] (0, 0) -- node [left,pos=1.1,xshift=5]{$z^C$}(0, 48);
	\draw[shift={(671.891, 159.941)}, rotate=18.4441, red, ->,line width = 1.2] (0, 0) -- node [right,pos=0.9]{$x^C$} (48, 0);
	\draw[dotted] (672, 160) -- (640, 448);
	\draw[dotted] (640, 448) -- (480, 192);
	\draw[->] (480, 192) -- node [left,pos=0.8]{ $t_z^{GP}$} (480, 448);

	\draw[red, ->, line width = 1.2] (256, 448) -- node [left,pos=1] {$z^G$} (256, 496);
	\draw[red, ->, line width = 1.2] (256, 448) -- node [below,pos=0.8]{$x^G$} (304, 448);
	\draw[red, ->, line width = 1.2] (480, 192) -- node [left,xshift=4,pos=1.2]{$z^P$} (480, 240);
	\draw[red, ->, line width = 1.2] (480, 192) -- node [right,pos=0.8,yshift=-2]{$x^P$} (528, 192);
	
\end{tikzpicture}}
}
\caption{The drawings show the definition of the three coordinate systems used throughout the article: the gel coordinate system (superscript $G$), the pinhole camera coordinate system (superscript $P$), and the real-world camera coordinate system (superscript $C$). In (a), an example of 3D displacement of a particle originally placed at $s_j$ is depicted. In (b), a pixel in the pinhole camera is mapped to the corresponding pixel in the real-world camera.}
\label{fig:frames_definition}
\end{figure}

A gel coordinate system (see Fig.~\ref{fig:frames_definition}(a)) was defined by placing the origin at one of the bottom corners of the layer containing the particles, the $z$ axis pointing towards the upper surface, and the $x$ and $y$ axes aligned with two of the horizontal edges. For each indentation step performed in simulation, the FEM provides the displacement field of the soft layer that comprises the particles. This displacement field is provided at the discrete nodes of the FEM mesh. For such nodes, also the initial position (at rest, before deformation) is known. In order to generate the dataset, a random distribution of particles was sampled for each indentation step, and an inverse distance weighted scheme\cite{idw_interpolation} was used to interpolate the displacement field at the corresponding particle location $s_j^G$, for $j = 0,...,N_\text{p}-1,$ where $N_\text{p}$ is the number of particles and the superscript $G$ indicates the gel coordinate system. The 3D displacement of the $j$-th particle is denoted in the following as $\Delta s_j^G$. The strategy followed was to project the particles to the image plane using an ideal pinhole camera model,\cite{cv_book} and only account for the camera's non-idealities at a later stage,\cite{sim2real_iros} as described in \mbox{Section \ref{subsec:test_set}}. Therefore, as depicted in Fig.~\ref{fig:frames_definition}, the position $s_j^G$ and the respective displacement $\Delta s_j^G$ were first transformed from the gel coordinate system to the 3D pinhole camera coordinate system (indicated by the superscript $P$) as
\begin{gather}
s_j^P  = R^{GP} s_j^G + t^{GP}, \label{eq:pos_projection1}\\
\Delta s_j^P = R^{GP} \Delta s_j^G
\end{gather}
where the rotation matrix $R^{GP}$ and the translation vector $t^{GP} := (t^{GP}_x,t^{GP}_y,t^{GP}_z)$ are the pinhole camera's extrinsic parameters. These parameters could be chosen arbitrarily, but they were actually chosen to be close to the real-world camera's extrinsic parameters, as discussed in \mbox{Section \ref{subsec:test_set}}. 
The pinhole image resolution was arbitrarily set to be 440$\times$440 pixels, and although the focal length could also be chosen arbitrarily in this step, in order to exactly capture the region where the particle layer (which has a square horizontal section of 30$\times$30 mm) is visible, this was set for both the image coordinates as 
\begin{align}
f := \dfrac{440}{30}t^{GP}_z.
\end{align}
The projection of the spherical particle centered at \mbox{$s_\text{p}^P$} via the pinhole camera model results in an ellipse on the image plane\cite{sphere_projection}. 
The derivation of the center, the axis lengths and the orientation of each ellipse can be found in Section 3 of the supplementary material. The ellipses can then be drawn using the drawing functionality of OpenCV\footnote{https://opencv.org/}. 

For each indentation step, one image at rest (projecting all the particles, i.e., by setting \mbox{$s_\text{p}^P=s_j^P$} for the $j$-th particle) and one image after deformation (setting \mbox{$s_\text{p}^P=s_j^P+\Delta s_j^P$} for the $j$-th particle) were generated. The images were initialized with black pixels, and each ellipse was drawn with a random RGB color to perturb the data with additional variability. The images were then converted to grayscale in a second step. An example of a simulated image is shown in Fig.~\ref{fig:teaser}(c). In order to further increase the training robustness, the number of the particles within the gel was slightly perturbed at each indentation step. Training features were then extracted from the images via two different methods:
\begin{enumerate}
	\item Optical flow features: For each indentation step, the dense optical flow between the image at rest and the image after deformation was computed using an algorithm based on Dense Inverse Search.\cite{dis_opticalflow} The \mbox{per-pixel} flow was then subsampled performing an average pooling in a grid of 88$\times$88 bins. The two Cartesian components of the optical flow resulted in two matrices, which were concatenated into a two-channel matrix. This method differs from previous work,\cite{sim2real_iros} where optical flow features were directly computed from the FEM displacement field, assuming that the density of the particles remained constant during an indentation. In reality, this is not the case for large indentations, as the particles tend to spread radially under pressure, and the method presented here can cope with such conditions.
	\item Raw features: The two images for each indentation step were subsampled to 88$\times$88 pixels and concatenated into a two-channel image, which was directly fed to the training algorithm.
\end{enumerate}
\subsubsection{Training labels}

\label{subsec:training_labels}
The same set of labels described in the following was assigned to each set of features to compose two separate training datasets. For each indentation step, the FEM simulations provide the three-dimensional contact force distribution at the surface nodes of the FEM mesh. Dividing the surface into a grid of 20$\times$20 bins,\cite{sferrazza_fem} the force components at the nodes falling inside a bin were summed to obtain a 20$\times$20 three-channel matrix, representing the training label for the corresponding indentation step datapoint. Examples of ground truth labels are shown in Fig.~\ref{fig:results}. In this work, a node was assigned to a certain bin depending on its initial position before deformation, in order to simplify the binning at the boundaries of the gel, which can vary with deformation. As an alternative, it would also be possible to assign the nodes to the bins according to the position after deformation, by introducing an adaptive binning strategy at the boundaries of the grid.
\subsubsection{Test dataset}
\label{subsec:test_set}
In order to evaluate the real-world performance of the models described in Section \ref{sec:learning}, 1100 test datapoints were collected in an experimental setup, using a programmable milling machine (Fehlmann PICOMAX 56 TOP) to make vertical and shear-dominant indentations with the six black indenters shown in Fig.~\ref{fig:indenters}, as well as multi-contact indentations with two spherically-ended indenters placed at different heights. The resulting test dataset induced total forces up to 4.5 N in the vertical direction and up to 3.8 N in each of the horizontal directions. These ranges differ from the training data ranges, which also included data inducing larger strains and where the material model fit was less accurate. Such higher-strain data showed improved generalization in the learning and for this reason were included in the training dataset.

\begin{figure}[H]
	\centering
	\subcaptionbox{Calibration setup}{
		\includegraphics[width=0.35\columnwidth]{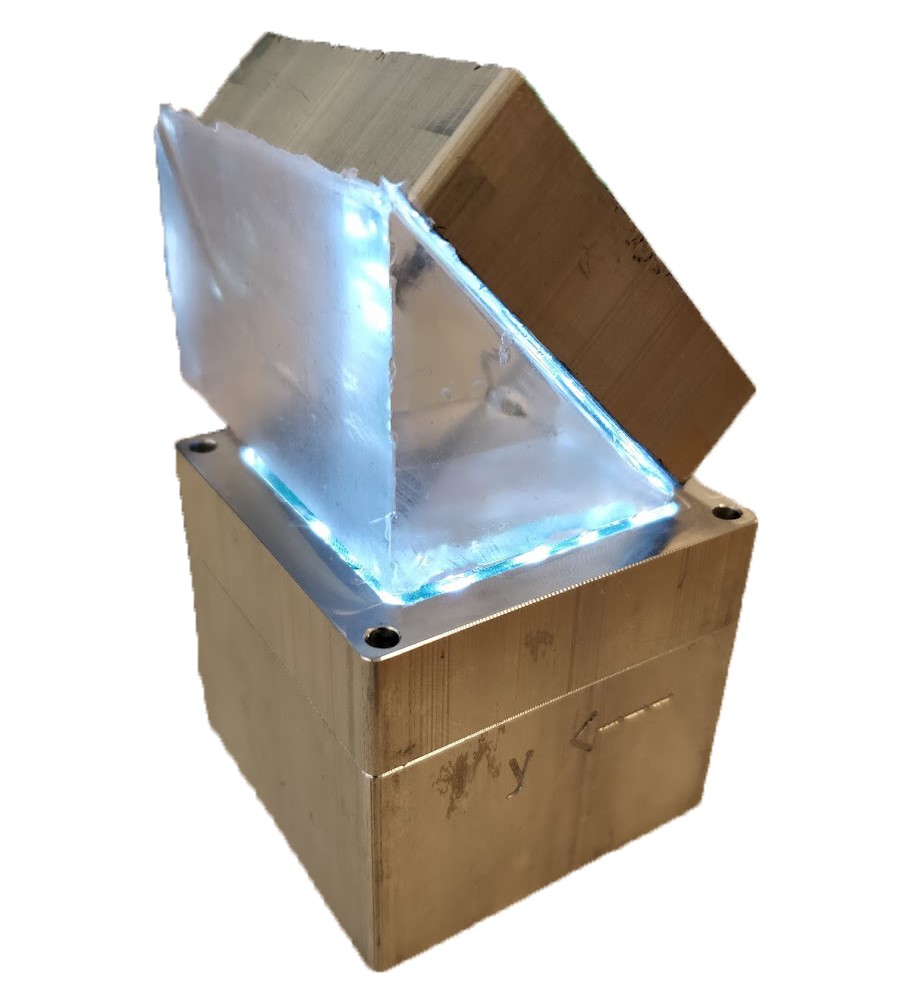}
	}
	\hfill
	\subcaptionbox{Sample calibration images}{
		\includegraphics[width=0.5\columnwidth]{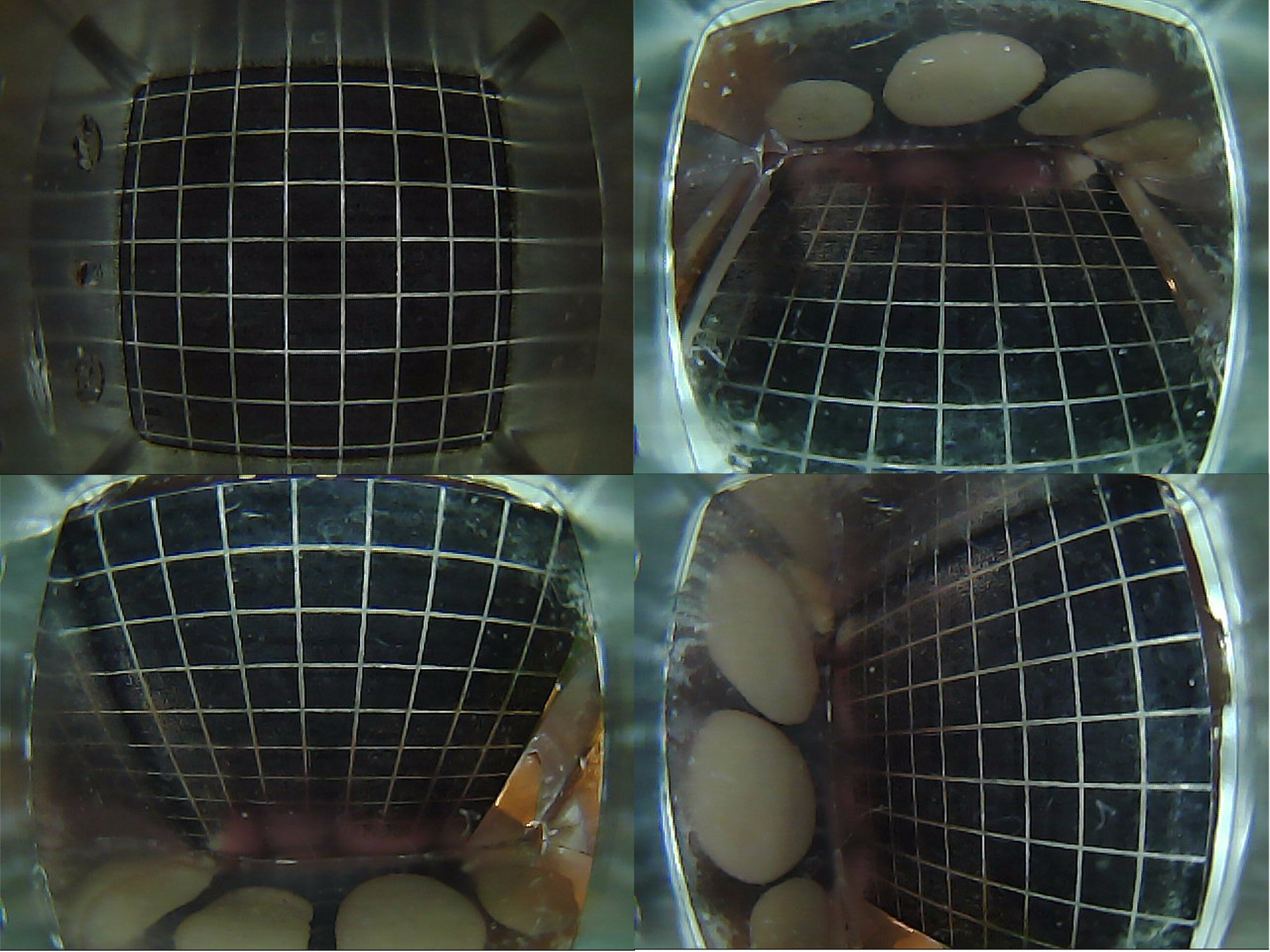}
	}
	\caption{The calibration images, examples of which are shown in (b), were shot through a silicone medium during fabrication, in order to account for the refraction index of the soft materials. As shown in (a), this was done straight after casting the first layer, by placing additional silicone parts between the first layer and a grid pattern attached to an aluminum surface.}
	\label{fig:calibration}
\end{figure}

\begin{figure}[H]
	\centering
	\subcaptionbox{Original image}{
		\includegraphics[height=0.4\columnwidth]{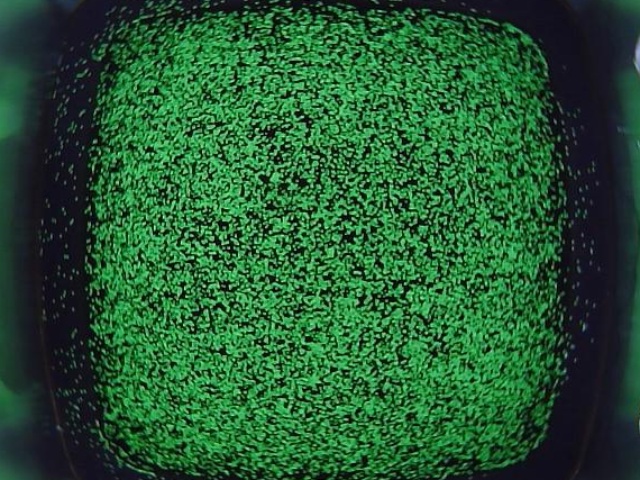}
	}
	\hfill
	\subcaptionbox{Image before feature extraction}{
		\includegraphics[height=0.4\columnwidth]{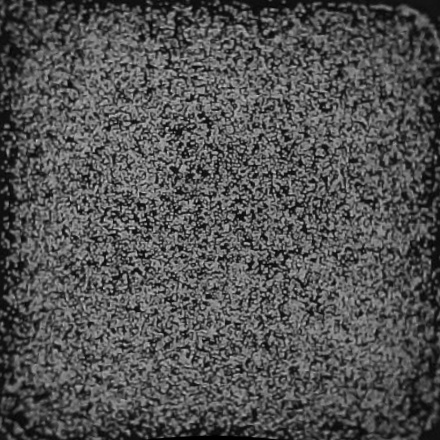}
	}	
	\caption{The original image taken from the real-world camera, shown in (a), was converted to grayscale and remapped as if it was taken from the ideal pinhole camera. A refinement procedure was applied to account for inaccuracies introduced during calibration. The resulting image in (b) shows the particle layer in its actual squared geometry, covering the entire image frame.}
	\label{fig:remapped_image}
\end{figure}

During the test data collection procedure, the images taken by the real-world camera were recorded. Since the models were trained with features obtained from images generated via a pinhole camera projection, a further procedure was needed to account for the camera's non-idealities on real-world images.\cite{sim2real_iros} This procedure is denoted as remapping and it essentially maps the pixels from a real-world image (converted to grayscale) to the pixels of an image of the same scene as if it was taken from the ideal pinhole camera used for the training dataset. The remapping procedure requires two main steps:
\begin{enumerate}
	\item Calibration step: during fabrication, seven images of a grid pattern were shot through a silicone medium, see Fig.~\ref{fig:calibration}. In this way, it is possible to account for the refractive index of the soft materials. Using a fisheye camera calibration toolbox,\cite{scaramuzza_toolbox} the images were used to obtain both the extrinsic parameters $R^{GC}$ and $t^{GC}$ of the real-world camera as well as a transformation function from the actual camera 3D coordinate system to the real-world image. The extraction of the extrinsic parameters was achieved by providing the calibration toolbox with a calibration image where the origin of the grid pattern coincided with the origin of the gel coordinate system.
	\item Interpolation step: for each pixel in the fictitious pinhole image, the corresponding pixel in the real-world image was obtained, via a procedure sketched in Fig.~\ref{fig:frames_definition}(b). For this step, the pixels were assumed to be placed approximately at a fixed $z$ coordinate in the pinhole camera coordinate system, set here with the bottom of the gel layer. The details of the interpolation procedure are further detailed in Section 4 of the supplementary material.
	
	As shown in Fig.~\ref{fig:frames_definition}(b), the approximation introduced above has a smaller effect when the pinhole extrinsic parameters $R^{GP}$ and $t^{GP}$ are close to the real-world camera extrinsic parameters $R^{GC}$ and $t^{GC}$, respectively. As mentioned in Section \ref{subsec:training_features}, since the pinhole extrinsic parameters can be set arbitrarily, these were indeed chosen to be close to the expected real-world extrinsic parameters to limit the impact of the approximation. While the calibration parameters are fixed across images of the same camera and can be computed offline, the interpolation step needs to be performed for each image.
\end{enumerate}
The extrinsic parameters obtained during calibration are very sensitive to the exact placement of the grid pattern for the corresponding calibration image. This requires pressing the grid pattern against the silicone medium just enough to remove the air in the middle without penetrating the soft material, which is challenging to achieve in reality. Therefore, a grid search (in the submillimeter range) was performed in the vicinity of the translation vector $t^{GC}$, in order to make the particles in a sample remapped image taken at rest match the entire image frame. For this, after a series of dilation and erosion steps, a bounding box around the pixels can be easily computed using OpenCV and compared to the frame boundaries. A refined, remapped image is shown in Fig.~\ref{fig:remapped_image}, where lens distortion effects and misalignments were successfully compensated for.

After remapping, the same image features described in Section \ref{subsec:training_features} were extracted from the images. Since no real-world sensor can provide ground truth contact force distributions, these were extracted in simulation as described in Section \ref{subsec:training_labels} and in previous work,\cite{sferrazza_fem} and assigned to the corresponding features to compose two test datasets. Note that since the real-world camera non-idealities can be compensated in the remapping step, which does not affect training, this enables the transfer of models trained on the pinhole data across multiple instances of fabricated sensors, provided that the camera calibration is performed as described above. The remapping procedure described here aims to compensate only for the camera mismatches and does not serve as a calibration for the FEM model, which was independently characterized in previous work,\cite{sferrazza_fem} as further detailed in the supplementary material.
	
%


\subsection{Learning architecture} \label{sec:learning}
The same learning architecture was employed for both training datasets, that is, on those containing optical flow features and raw features, respectively. The architecture consists of a convolutional neural network, designed as a lightweight version of u-net,\cite{unet_paper} and tailored to the estimation of the force distribution from tactile features. In fact, this estimation problem can be formulated as an image-to-image translation\cite{i2i_translation} (known also as pixel-wise regression). A sketch of the architecture is shown in Fig.~\ref{fig:unet}. The neural network exhibits an encoder-decoder structure, where feature information is first increased in the contraction step by doubling the channels between each pooling operation. In the decoding step, the force distribution is then computed through upconvolutions and concatenations of high-resolution features extracted during the contraction step. As a result, the architecture has the effect of both capturing context and enabling precise localization.

\begin{figure*}
	\centering
	\subcaptionbox{The learning architecture}{
		\scalebox{0.85}{\definecolor{darkgray}{rgb}{0.663,0.663,0.663}%
\definecolor{customgreen}{rgb}{0.796,0.890,0.824}%
\definecolor{verylightgray}{rgb}{0.927,0.927,0.927}%

\begin{tikzpicture}[scale=0.5,thick,x=1mm,y=1mm]
\node [draw, line width = 1.5,rounded corners, minimum width = 15, minimum height = 160, fill = customgreen] (input){\rotatebox{90}{input features}};
\node [anchor=south west, draw, line width = 1.5,rounded corners, minimum width = 20, minimum height = 160, fill = darkgray] (down1)at ([xshift=30]input.south east){\rotatebox{90}{encoder, 8}};
\node [anchor=south west,draw, line width = 1.5,rounded corners, minimum width = 30, minimum height = 140, fill = darkgray] (down2)at ([xshift=30]down1.south east){\rotatebox{90}{encoder, 16}};
\node [anchor=south west,draw, line width = 1.5,rounded corners, minimum width = 40, minimum height = 120, fill = darkgray] (down3)at ([xshift=30]down2.south east){\rotatebox{90}{encoder, 32}};
\node [anchor=south west,draw, line width = 1.5,rounded corners, minimum width = 50, minimum height = 100, fill = darkgray] (down4)at ([xshift=30]down3.south east){\rotatebox{90}{encoder, 64}};
\node [anchor=south west,draw, line width = 1.5,rounded corners, minimum width = 60, minimum height = 100, fill = gray] (bottleneck1)at ([xshift=30]down4.south east){\rotatebox{90}{3$\times$3 conv, 128}};
\node [anchor=south west,draw, line width = 1.5,rounded corners, minimum width = 60, minimum height = 100, fill = gray] (bottleneck2)at ([xshift=30]bottleneck1.south east){\rotatebox{90}{3$\times$3 conv, 128}};
\node [anchor=south west,draw, line width = 1.5,rounded corners, minimum width = 50, minimum height = 100, fill = darkgray] (up1)at ([xshift=30]bottleneck2.south east){\rotatebox{90}{decoder, 64}};
\node [anchor=south west,draw, line width = 1.5,rounded corners, minimum width = 40, minimum height = 120, fill = darkgray] (up2)at ([xshift=30]up1.south east){\rotatebox{90}{decoder, 32}};
\node [anchor=south west,draw, line width = 1.5,rounded corners, minimum width = 15, minimum height = 110, fill = white] (final)at ([xshift=30]up2.south east){\rotatebox{90}{3$\times$3 conv, 3, 0p}};
\node [anchor=south west,draw, line width = 1.5,rounded corners, minimum width = 15, minimum height = 110, fill = customgreen] (force)at ([xshift=30]final.south east){\rotatebox{90}{force distribution}};

\draw [->] (input.east|-down1.west) -- (down1.west);
\draw [->] (down1.east|-down2.west) -- (down2.west);
\draw [->] (down2.east|-down3.west) -- (down3.west);
\draw [->] (down3.east|-down4.west) -- (down4.west);
\draw [->] (down4.east|-bottleneck1.west) -- (bottleneck1.west);
\draw [->] (bottleneck1.east|-up1.west) -- (bottleneck2.west);
\draw [->] (bottleneck2.east|-up1.west) -- (up1.west);
\draw [->] (up1.east|-up2.west) -- (up2.west);
\draw [->] (up2.east|-final.west) -- (final.west);
\draw [->] (final.east|-force.west) -- (force.west);
\draw [->,dotted,line width=1.5] ([yshift=-10]down3.north east|-up2.north west) -- ([yshift=-10]up2.north west);
\draw [->,dotted,line width=1.5] (down4.south) to [out=-30,in=-150] (up1.south);

\end{tikzpicture}}
	}
	\vskip \baselineskip
	\subcaptionbox{An encoder block}{
		\scalebox{0.85}{\definecolor{darkgray}{rgb}{0.663,0.663,0.663}%
\definecolor{customgreen}{rgb}{0.796,0.890,0.824}%
\definecolor{verylightgray}{rgb}{0.927,0.927,0.927}%

\begin{tikzpicture}[scale=0.5,thick,x=1mm,y=1mm]
\node [draw, line width = 1.5,rounded corners, minimum width = 20, minimum height = 160, fill = customgreen] (input){\rotatebox{90}{previous output, $c$}};
\node [anchor=south west, draw, line width = 1.5,rounded corners, minimum width = 40, minimum height = 160, fill = gray] (down1)at ([xshift=30]input.south east){\rotatebox{90}{3$\times$3 conv, 2$c$}};
\node [anchor=south west,draw, line width = 1.5,rounded corners, minimum width = 40, minimum height = 160, fill = gray] (down2)at ([xshift=30]down1.south east){\rotatebox{90}{3$\times$3 conv, 2$c$}};
\node [anchor=south west,draw, line width = 1.5,rounded corners, minimum width = 40, minimum height = 140, fill = verylightgray] (down3)at ([xshift=30]down2.south east){\rotatebox{90}{1/2 max pooling}};

\draw [->] (input.east|-down1.west) -- (down1.west);
\draw [->] (down1.east|-down2.west) -- (down2.west);
\draw [->] (down2.east|-down3.west) -- (down3.west);
\draw [->] (down3.east) -- ([xshift=30]down3.east);

\end{tikzpicture}}
	}
	\hfill
	\subcaptionbox{A decoder block}{
		\scalebox{0.85}{\definecolor{darkgray}{rgb}{0.663,0.663,0.663}%
\definecolor{customgreen}{rgb}{0.796,0.890,0.824}%
\definecolor{verylightgray}{rgb}{0.927,0.927,0.927}%

\begin{tikzpicture}[scale=0.5,thick,x=1mm,y=1mm]
\node [draw, line width = 1.5,rounded corners, minimum width = 40, minimum height = 140, fill = customgreen] (input){\rotatebox{90}{previous output, $2c$}};
\node [anchor=south west, draw, line width = 1.5,rounded corners, minimum width = 20, minimum height = 160, fill = verylightgray] (down1)at ([xshift=30]input.south east){\rotatebox{90}{3$\times$3 upconv, $c$}};
\node [anchor=south west,draw, line width = 1.5,rounded corners, minimum width = 20, minimum height = 160, fill = gray] (down2)at ([xshift=30]down1.south east){\rotatebox{90}{3$\times$3 conv, $c$}};
\node [anchor=south west,draw, line width = 1.5,rounded corners, minimum width = 20, minimum height = 160, fill = gray] (down3)at ([xshift=30]down2.south east){\rotatebox{90}{3$\times$3 conv, $c$}};

\draw [->] (input.east|-down1.west) -- (down1.west);
\draw [->] (down1.east|-down2.west) -- (down2.west);
\draw [->] (down2.east) -- (down3.west);
\draw [->] (down3.east) -- ([xshift=30]down3.east);

\end{tikzpicture}}
	}
	\caption{In (a), a diagram of the learning architecture is shown. The encoder and decoder blocks are summarized in (b) and (c), respectively. All the blocks in green serve as placeholders. ``3$\times$3 conv, $c$'' indicates a convolutional layer with a 3$\times$3 filter size and $c$ output channels, while ``3$\times$3'' upconv, $c$” indicates an upconvolution that doubles the input size. The dotted lines (omitted in (b) and (c)) indicate the concatenation of an earlier layer output with upsampled information. After each convolutional layer, with the exception of the white one before the final output, batch normalization and rectified linear units were employed. “0p” indicates no padding. Where not indicated, all convolutional filters have unit zero-padding and unit stride.}
	\label{fig:unet}
\end{figure*}
\section{Results} \label{sec:results}
The learning architecture was trained twice from scratch using I) the training dataset comprising averaged optical flow features and discretized force distribution labels, and II) the training dataset comprising raw image features and discretized force distribution labels. Both datasets were generated entirely in simulation and both sets of training features contained two-channel 88$\times$88 matrices (or images), as described in Section \ref{subsec:training_features}. The architecture was trained with the AdamW optimizer\cite{adamw_paper} by minimizing a mean-squared loss (normalized by the maximum value per channel) with a learning rate of 1e-3 and a batch size of 256. During training, the datasets were randomly augmented by appropriately flipping the features and labels, exploiting the symmetry of the gel geometry and the pinhole camera projection. For the raw-feature dataset, the images were additionally augmented by perturbing the image brightness and adding salt-and-pepper noise.

After training in PyTorch\footnote{https://pytorch.org/}, the models were converted to the ONNX format, and used in real-time via the ONNX Runtime framework\footnote{https://www.onnxruntime.ai/}. This generally led to a 4x inference speed-up on the CPU of a standard laptop (dual-core, 2.80 GHz), compared to the inference in PyTorch. 

The performance of both trained models was evaluated on the corresponding synthetic validation datasets, picked randomly as the 20\% of the indentation trajectories in the appropriate training dataset. Additionally, the models were evaluated on the corresponding real-world test dataset described in Section \ref{subsec:test_set}. Table \ref{table:errors} summarizes the results based on two different error metrics for each force component: I) RMSE, that is the root-mean-squared error on the respective component of the force distribution, II) RMSET, that is the root-mean squared error on the respective component of the total force, which was obtained by summing the force distribution over all the bins. The range of total forces in the corresponding dataset is also shown in the table. In addition, Table \ref{table:absolute_errors} reports the mean and the standard deviation of the absolute errors, for the bin-wise and total force predictions on the real-world test data. 

\begin{table*}[]
	\centering
	\begin{tabular}{lccccccccc}
		\hline
		& \multicolumn{3}{c}{RMSE [N]}          & \multicolumn{3}{c}{RMSET [N]}                     & \multicolumn{3}{c}{Range of total forces   [N]}          \\ \hline
		& $x$      & $y$      & $z$      & $x$     & $y$        & $z$       & $x$          & $y$               & $z$ \\ \hline
		Optical-flow (sim)  & 0.006  & 0.006  & 0.013  & 0.187 & 0.164    & 0.577   & -5.0 -- 5.0 & -5.0 -- 5.0      & -16.0 -- 0 \\
		Raw-feature (sim)   & 0.006  & 0.005  & 0.012  & 0.120 & 0.132    & 0.314   & -5.0 -- 5.0 & -5.0 -- 5.0      & -16.0 -- 0 \\
		Optical-flow (real) & 0.006  & 0.007  & 0.018  & 0.190 & 0.230    & 0.914   & -3.2 -- 3.2 & -3.8 -- 3.8      & -4.5 -- 0 \\
		Raw-feature (real)  & 0.005  & 0.007  & 0.014  & 0.267 & 0.296    & 0.362   & -3.2 -- 3.2 & -3.8 -- 3.8      & -4.5 -- 0      \\
		\hline
	\end{tabular}
\caption{The table shows the error metrics of the trained models on the validation datasets extracted in simulation and the test datasets collected in reality, for both the cases where optical flow features and raw features were used as inputs.}
\label{table:errors}
\end{table*}

\begin{table*}[]
	\centering
	\begin{tabular}{lcccccc}
		\hline
		& \multicolumn{3}{c}{MAE [N]} & \multicolumn{3}{c}{SDAE [N]} \\ \hline
		& $x$     & $y$     & $z$     & $x$      & $y$     & $z$     \\ \hline
		Optical-flow (bin)   & 0.001   & 0.001   & 0.004   & 0.006    & 0.007   & 0.018   \\
		Raw-feature (bin)    & 0.001   & 0.001   & 0.003   & 0.005    & 0.007   & 0.014   \\
		Optical-flow (total) & 0.103   & 0.110   & 0.645   & 0.159    & 0.202   & 0.648   \\
		Raw-feature (total)  & 0.099   & 0.107   & 0.238   & 0.248    & 0.275   & 0.273   \\ \hline
	\end{tabular}
\caption{The table shows additional error metrics on the real-world test sets in terms of the absolute errors for bin-wise and total force predictions, namely the mean absolute error (MAE), and the standard deviation of the absolute errors (SDAE).}
\label{table:absolute_errors}
\end{table*}

\begin{figure*}[!ht]
	\centering
	\includegraphics[width=0.9\textwidth]{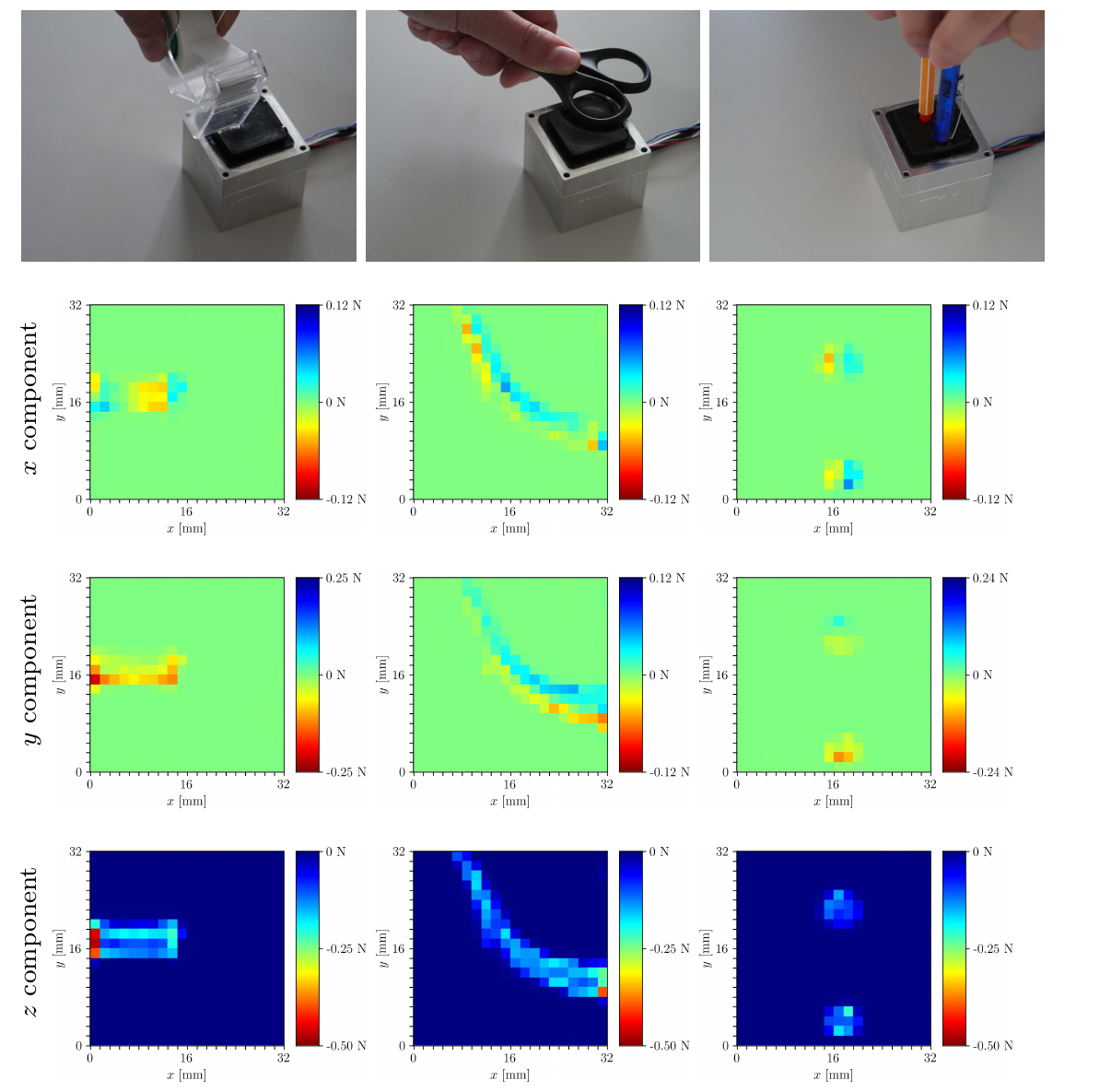}
	\caption{The figures show sensible predictions for the different contact conditions that are shown in the first row. The $x$, $y$, and $z$ components of the predicted force distributions are shown in the second, third, and fourth rows, respectively. In the first column, the tape dispenser was initially pressed against the gel and then translated to induce higher shear forces in the negative $y$ direction. In the second column, the contact with an object that differs significantly from those in the training set is shown, while the third column shows the contact with multiple bodies (not included in the training data), the lower of which is laterally translated as shown by the asymmetrical shear component in the $y$ direction.}
	\label{fig:generic_indentations}
\end{figure*}

As the numerical results indicate, there is a slight difference in accuracy between the horizontal and vertical components of the predictions. This may be explained by the fact that during the vertical indentations, the shear forces were rather small or close to zero. More importantly, the raw-feature model outperformed the optical-flow model in most of the metrics on the corresponding real-world test dataset. In fact, while in practice the location accuracy for both models was similar, the optical-flow features tend to mitigate the differences across indenters under real-world noise, therefore resulting in inaccurate force predictions. On the other hand, overall the raw-feature model showed a better transfer from simulation to reality, especially retaining a considerably higher accuracy in the vertical component. In addition to the difference in accuracy, the model trained on the raw image features does not require the extraction of the optical flow, which was the bottleneck for the model trained on optical flow features. As a result, since the model inference only takes about two milliseconds, the whole raw-feature pipeline (including the image acquisition and remapping) runs in real-time at 120 Hz on CPU, compared to the 50 Hz of the optical-flow pipeline.

The real-time performance of the raw-feature pipeline is shown in the supplementary video, where contact conditions with arbitrary objects were explored (see Fig.~\ref{fig:generic_indentations}). The estimation of the force distributions on samples in the test set with the model trained on raw features are shown in Fig.~\ref{fig:results}. Additional results and comparisons are available in Section 5 of the supplementary material.
\begin{figure*}
	\centering
	\includegraphics[width=0.9\textwidth]{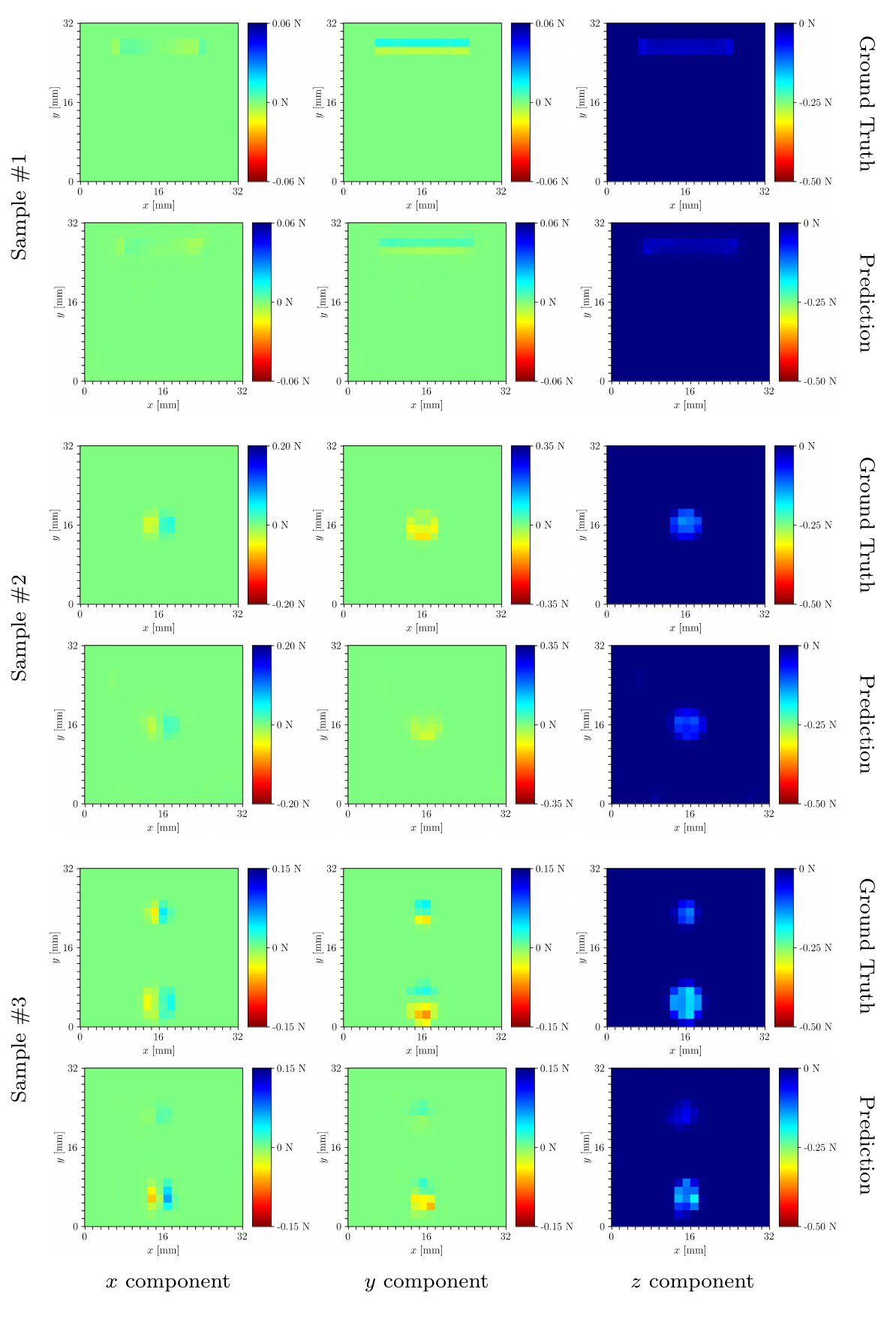}
	\caption{The figures show the ground truth (first, third, and fifth rows) and predicted (second, fourth, and sixth rows) force distribution components ($x$ in the first column, $y$ in the second column, and $z$ in the third column) for different samples and indenters in the real-world test dataset. Predictions were made with the raw-feature model. The first two rows show vertical indentation; the third and fourth rows show a shear-dominant indentation, and the last two rows a multi-contact indentation.}
	\label{fig:results}
\end{figure*}


\section{Conclusion} \label{sec:conclusion}
This work has discussed strategies to simulate the images captured by a vision-based tactile sensor. Starting from FEM simulations, the displacement field was processed to generate training features for a supervised learning architecture that mapped these features to contact force distribution labels. The resulting models are directly transferable across multiple instances of real-world sensors, since the training procedure does not make use of real-world images. Two different strategies were compared, with the model obtained from raw features outperforming a model based on optical flow features for both real-world accuracy and inference speed. In addition to providing a physical quantity directly interpretable across robotic tasks, the extraction of accurate force distributions also provides an abstraction from the image pixels that bypasses the remaining mismatch between real and simulated images.

Since this work aimed to provide a comparison between the two approaches, the same input and output sizes were employed for both strategies. However, given the gain in prediction speed, the raw-feature approach may be extended to use higher-resolution features or to predict the force distribution on a finer grid by trading off the sensing frequency. As shown in Table \ref{table:errors}, a gap still remains between simulation and reality, which could be addressed by explicitly addressing the domain transfer problem. This issue will be the subject of future work. 

The simulator described in this work provides highly accurate force distribution labels to train learning-based models suitable for real-time inference. However, the simulator itself is not running in real-time, due to the computational complexity of the finite element method. While this was not in the scope of this work, different simulation techniques can trade off accuracy to achieve real-time capabilities and become suitable to warm-start the training of tactile policies in simulation, as detailed in a related work.\cite{zeroshot_thomas}


\section*{Acknowledgements} 
The authors would like to thank Michael Egli and Matthias Mueller for their support in the sensor manufacture, and Thomas Bi for the discussions on the sensing pipeline.

\section*{Author Disclosure Statement}
No competing financial interests exist.


\bibliographystyle{ieeetr}
\bibliography{references.bib}

\end{multicols}
\end{document}


\title{Sim-to-real for high-resolution optical tactile sensing:\\ From images to 3D contact force distributions \\ - \\ Supplementary Material}
	\def\RunningHead{
		Sim-to-real for high-resolution optical tactile sensing
	}
	\def\RunningAuthor{Sferrazza and D'Andrea}
	\author{Carmelo Sferrazza and Raffaello D'Andrea
		\thanks{Institute for Dynamic Systems and Control, ETH Zurich, 8092 Zurich, Switzerland. \newline\hspace*{6mm} Correspondence to:
			{\tt\small csferrazza@ethz.ch}}%
	}
	
	\maketitle


	\begin{multicols}{2}
		
\section{Fabrication}
\begin{figure*}[t]
	\centering
	\subcaptionbox{Exploded view of the three-layer structure}{
		\includegraphics[height=0.6\columnwidth]{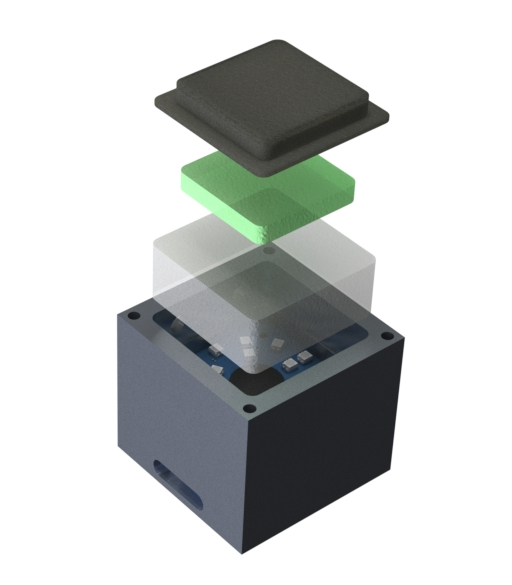}
	}
	\subcaptionbox{Mold assembled before casting the first layer}{
		\includegraphics[height=0.6\columnwidth]{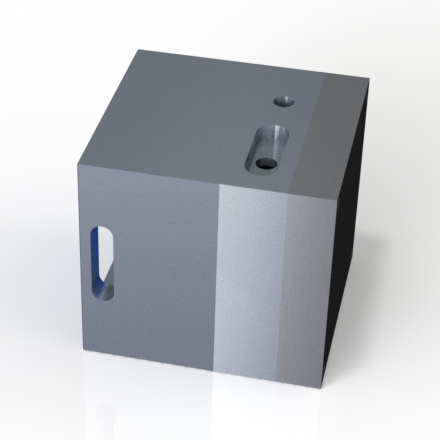}
	}
	\vskip \baselineskip
	\subcaptionbox{Cross-section after casting the \\first layer}{
		\includegraphics[width=0.6\columnwidth]{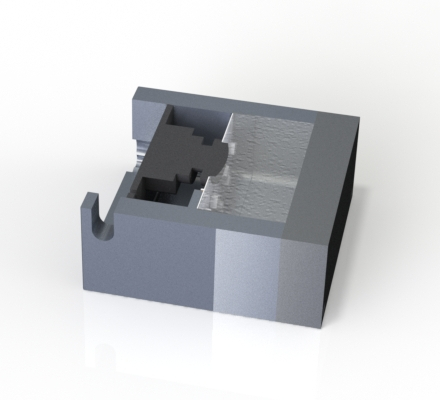}
	} \hfill
	\subcaptionbox{Cross-section after casting the second layer}{
		\includegraphics[width=0.6\columnwidth]{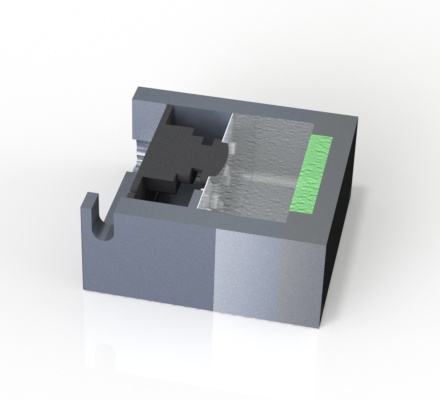}
	}	\hfill
	\subcaptionbox{Cross-section after casting the third layer}{
		\includegraphics[width=0.6\columnwidth]{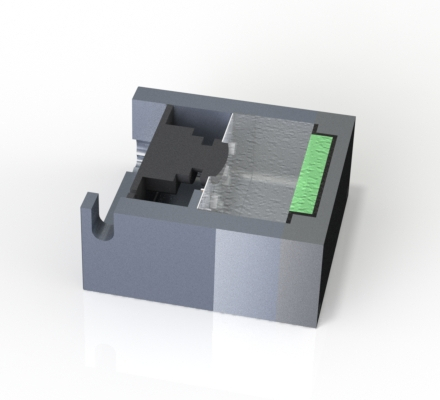}
	} 
	\caption{The figure details the sensor's fabrication. The soft materials are arranged in a three-layer structure (see (a)) on top of the camera and the LEDs, and were poured into the mold from the side, through lateral cavities such as those shown in (b). Three different lids (see (c)-(e)) were employed for each of the soft layers.}
	\label{fig:fabrication}
\end{figure*}

The soft materials are arranged in three layers, as shown in Fig.~\ref{fig:fabrication}(a), and were poured (after degassing in a vacuum chamber) on top of the camera \mbox{(ELP USBFHD06H} with a fisheye lens) and the surrounding LEDs with the mold lying on one side, i.e., with the camera pointing sideways, as shown in Fig.~\ref{fig:fabrication}(b). The fabrication strategy followed previous work \cite{sferrazza_sensors}, but it is presented extensively here to provide additional details for reproducibility:
\begin{enumerate}
	\item A first layer of Elastosil\textregistered{} RT 601 RTV-2 (mixing ratio 7:1, shore hardness 45A) was poured into the mold, closed with a first lid (see Fig.~\ref{fig:fabrication}(c)). The mold was then placed into an oven at \mbox{80 \textdegree C} for 20 minutes for curing. This layer serves as a stiff base and facilitates light diffusion.
	\item A release agent (Mann Ease Release\textsuperscript{TM} 200) was sprayed before assembling the second lid, shown in Fig.~\ref{fig:fabrication}(d). Then, a layer of Ecoflex\textsuperscript{TM} GEL (mixing ratio 1:1, very soft, with shore hardness 000-35), mixed with green, fluorescent spherical particles (with a diameter of 150 to 180 $\mu$m), was poured into the mold. The mold was finally placed into an oven at 80 \textdegree C for 20 minutes for curing.
	\item Finally, a layer of Elastosil\textregistered{} RT 601 \mbox{RTV-2} (mixing ratio 25:1, shore hardness 10A), mixed with black silicone color (Elastosil\textregistered{} Color Paste FL), was poured into the mold, closed with a third lid, shown in Fig.~\ref{fig:fabrication}(e). The mold was then placed into an oven at \mbox{80 \textdegree C} for 45 minutes for curing. This layer is stiffer than the Ecoflex GEL, and shields the sensor from damage and light disturbances.
	\item After removing the last lid, the sensor was placed back in the oven at 60 \textdegree C for 8 hours. This step has been shown to reduce stiffening caused by the aging of the materials\cite{sferrazza_fem}.
\end{enumerate}

The two soft upper layers amount to a rectangular prism of 32$\times$32$\times$6 mm. 

\section{The FEM simulation environment}
The FEM simulation environment was created in \mbox{Abaqus/Standard\cite{abaqus_manual}} following previous \mbox{work\cite{sferrazza_fem}}, where the Ecoflex GEL and the black Elastosil layer have both been characterized as hyperelastic materials using second-order Ogden \mbox{models\cite{ogden_model}}. Given the large difference in hardness, the stiff base layer was considered \mbox{rigid\cite{sferrazza_fem}} in the FEM simulations discussed in the article. The contact between the top surface and the indenters was modeled as a hard contact and discretized with a surface-to-surface method. The basic Coulomb friction model available in Abaqus was employed, where the friction coefficient was assumed to be constant and was used as a tuning parameter, as described in the following.

\begin{figure*}
	\begin{subfigure}{0.49\textwidth}
		\hspace{1.75mm}
		\scalebox{0.8}{
\begin{tikzpicture}

\definecolor{color0}{rgb}{0.12156862745098,0.466666666666667,0.705882352941177}

\begin{axis}[
legend cell align={left},
legend style={
	fill opacity=0.8,
	draw opacity=1,
	text opacity=1,
	at={(0.97,0.13)},
	anchor=south east,
	draw=white!80!black
},
tick align=outside,
tick pos=left,
x grid style={white!69.0196078431373!black},
xlabel={Horizontal displacement [mm]},
xmin=-0.1, xmax=3.15,
xtick style={color=black},
xlabel style={yshift=-2pt},
xtick={-0.5,0,0.5,1,1.5,2,2.5,3,3.5},
xticklabels={\ensuremath{-}0.5,0.0,0.5,1.0,1.5,2.0,2.5,3.0,3.5},
y grid style={white!69.0196078431373!black},
ylabel={Horizontal force [N]},
ymin=-0.05, ymax=1.3,
ytick style={color=black},
ytick={-0.2,0,0.2,0.4,0.6,0.8,1,1.2,1.4},
yticklabels={\ensuremath{-}0.2,0.0,0.2,0.4,0.6,0.8,1.0,1.2,1.4}
]
\path [draw=blue, semithick]
(axis cs:0,0.0210971079178223)
--(axis cs:0,0.0210971079178223);

\path [draw=blue, semithick]
(axis cs:0.5,0.47671553893771)
--(axis cs:0.5,0.47671553893771);

\path [draw=blue, semithick]
(axis cs:1,0.875566023740994)
--(axis cs:1,0.875566023740994);

\path [draw=blue, semithick]
(axis cs:1.5,1.15457266043193)
--(axis cs:1.5,1.15457266043193);

\path [draw=blue, semithick]
(axis cs:2,1.20693434503246)
--(axis cs:2,1.20693434503246);

\path [draw=blue, semithick]
(axis cs:2.5,1.22510429479324)
--(axis cs:2.5,1.22510429479324);

\path [draw=blue, semithick]
(axis cs:3,1.2279324618827)
--(axis cs:3,1.2279324618827);

\addplot [semithick, red, mark=asterisk, mark size=3, mark options={solid}, only marks]
table {%
0 -0.00014168
0.5 0.43007
1 0.82161
1.5 1.1338
2 1.1932
2.5 1.2014
3 1.2102
};
\addlegendentry{FEM}
\addplot [semithick, color0, mark=asterisk, mark size=3, mark options={solid}, only marks]
table {%
0 0.0210971079178223
0.5 0.47671553893771
1 0.875566023740994
1.5 1.15457266043193
2 1.20693434503246
2.5 1.22510429479324
3 1.2279324618827
};
\addlegendentry{F/T sensor}
\node[label={RMSE = 0.022 N}] at (axis cs:2.35,-0.07) {};
\end{axis}

\end{tikzpicture}}
		\vspace{-2mm}
		\caption{Shear with spherical indenter (3DP)}
	\end{subfigure}
	\begin{subfigure}{0.49\textwidth}
		\hspace{3.5mm}
		\scalebox{0.8}{
\begin{tikzpicture}

\definecolor{color0}{rgb}{0.12156862745098,0.466666666666667,0.705882352941177}

\begin{axis}[
legend cell align={left},
legend style={
  fill opacity=0.8,
  draw opacity=1,
  text opacity=1,
  at={(0.635,0.33)},
  anchor=north west,
  draw=white!80!black
},
tick align=outside,
tick pos=left,
x grid style={white!69.0196078431373!black},
xlabel={Horizontal displacement [mm]},
xmin=-0.1, xmax=3.15,
xtick style={color=black},
xtick={-0.5,0,0.5,1,1.5,2,2.5,3,3.5},
xlabel style={yshift=-2pt},
xticklabels={\ensuremath{-}0.5,0.0,0.5,1.0,1.5,2.0,2.5,3.0,3.5},
y grid style={white!69.0196078431373!black},
ylabel={Horizontal force [N]},
ymin=-0.05, ymax=1.3,
ytick style={color=black},
ytick={-0.2,0,0.2,0.4,0.6,0.8,1,1.2,1.4},
yticklabels={\ensuremath{-}0.2,0.0,0.2,0.4,0.6,0.8,1.0,1.2,1.4}
]
\path [draw=blue, semithick]
(axis cs:0,0.0275648208986334)
--(axis cs:0,0.0275648208986334);

\path [draw=blue, semithick]
(axis cs:0.5,0.499017780965668)
--(axis cs:0.5,0.499017780965668);

\path [draw=blue, semithick]
(axis cs:1,0.912239281738971)
--(axis cs:1,0.912239281738971);

\path [draw=blue, semithick]
(axis cs:1.5,1.01493822299549)
--(axis cs:1.5,1.01493822299549);

\path [draw=blue, semithick]
(axis cs:2,1.03081953707362)
--(axis cs:2,1.03081953707362);

\path [draw=blue, semithick]
(axis cs:2.5,1.03059414083085)
--(axis cs:2.5,1.03059414083085);

\path [draw=blue, semithick]
(axis cs:3,1.033663750338)
--(axis cs:3,1.033663750338);

\addplot [semithick, red, mark=asterisk, mark size=3, mark options={solid}, only marks]
table {%
0 -0.00014168
0.5 0.43007
1 0.82161
1.5 1.1338
2 1.1932
2.5 1.2014
3 1.2102
};
\addlegendentry{FEM}
\addplot [semithick, color0, mark=asterisk, mark size=3, mark options={solid}, only marks]
table {%
0 0.0275648208986334
0.5 0.499017780965668
1 0.912239281738971
1.5 1.01493822299549
2 1.03081953707362
2.5 1.03059414083085
3 1.033663750338
};
\addlegendentry{F/T sensor}
\node[label={RMSE = 0.128 N}] at (axis cs:2.35,-0.07) {};
\end{axis}

\end{tikzpicture}}
				\vspace{-2mm}
		\caption{Shear with spherical indenter (SS)}
	\end{subfigure}
	\vskip \baselineskip
	\vspace{-2mm}
	\begin{subfigure}{0.49\textwidth}
				\hspace{1.75mm}
		\scalebox{0.8}{
\begin{tikzpicture}

\definecolor{color0}{rgb}{0.12156862745098,0.466666666666667,0.705882352941177}
\begin{axis}[
legend cell align={left},
legend style={
  fill opacity=0.8,
  draw opacity=1,
  text opacity=1,
  at={(0.635,0.33)},
  anchor=north west,
  draw=white!80!black
},
tick align=outside,
tick pos=left,
x grid style={white!69.0196078431373!black},
xlabel={Horizontal displacement [mm]},
xmin=-0.1, xmax=3.15,
xtick style={color=black},
xtick={-0.5,0,0.5,1,1.5,2,2.5,3,3.5},
xlabel style={yshift=-2pt},
xticklabels={\ensuremath{-}0.5,0.0,0.5,1.0,1.5,2.0,2.5,3.0,3.5},
y grid style={white!69.0196078431373!black},
ylabel={Horizontal force [N]},
ymin=-0.05, ymax=1.3,
ytick style={color=black},
ytick={-0.2,0,0.2,0.4,0.6,0.8,1,1.2,1.4},
yticklabels={\ensuremath{-}0.2,0.0,0.2,0.4,0.6,0.8,1.0,1.2,1.4}
]
\path [draw=blue, semithick]
(axis cs:0,0.00636435842669727)
--(axis cs:0,0.00636435842669727);

\path [draw=blue, semithick]
(axis cs:0.5,0.447015708462248)
--(axis cs:0.5,0.447015708462248);

\path [draw=blue, semithick]
(axis cs:1,0.701404705919901)
--(axis cs:1,0.701404705919901);

\path [draw=blue, semithick]
(axis cs:1.5,0.697143617576883)
--(axis cs:1.5,0.697143617576883);

\path [draw=blue, semithick]
(axis cs:2,0.707300015018)
--(axis cs:2,0.707300015018);

\path [draw=blue, semithick]
(axis cs:2.5,0.691638898847508)
--(axis cs:2.5,0.691638898847508);

\path [draw=blue, semithick]
(axis cs:3,0.700340818738262)
--(axis cs:3,0.700340818738262);

\addplot [semithick, red, mark=asterisk, mark size=3, mark options={solid}, only marks]
table {%
0 0.00033578
0.5 0.41013
1 0.73839
1.5 0.74419
2 0.76104
2.5 0.7637
3 0.7771
};
\addlegendentry{FEM}
\addplot [semithick, color0, mark=asterisk, mark size=3, mark options={solid}, only marks]
table {%
0 0.00636435842669727
0.5 0.447015708462248
1 0.701404705919901
1.5 0.697143617576883
2 0.707300015018
2.5 0.691638898847508
3 0.700340818738262
};
\addlegendentry{F/T sensor}
\node[label={RMSE = 0.037 N}] at (axis cs:2.35,-0.07) {};
\end{axis}
\end{tikzpicture}}
				\vspace{-2mm}
		\caption{Shear with square flat indenter (3DP)}
	\end{subfigure}
	\begin{subfigure}{0.49\textwidth}
		\hspace{3.5mm}
		\scalebox{0.8}{
\begin{tikzpicture}

\definecolor{color0}{rgb}{0.12156862745098,0.466666666666667,0.705882352941177}

\begin{axis}[
legend cell align={left},
legend style={
  fill opacity=0.8,
  draw opacity=1,
  text opacity=1,
  at={(0.635,0.33)},
  anchor=north west,
  draw=white!80!black
},
tick align=outside,
tick pos=left,
x grid style={white!69.0196078431373!black},
xlabel={Horizontal displacement [mm]},
xmin=-0.1, xmax=3.15,
xtick style={color=black},
xtick={-0.5,0,0.5,1,1.5,2,2.5,3,3.5},
xticklabels={\ensuremath{-}0.5,0.0,0.5,1.0,1.5,2.0,2.5,3.0,3.5},
y grid style={white!69.0196078431373!black},
ylabel={Horizontal force [N]},
ymin=-0.14554322, ymax=3.24480682,
ytick style={color=black},
ytick={-0.5,0,0.5,1,1.5,2,2.5,3,3.5},
yticklabels={\ensuremath{-}0.5,0.0,0.5,1.0,1.5,2.0,2.5,3.0,3.5}
]
\path [draw=blue, semithick]
(axis cs:0,0.0274873742456871)
--(axis cs:0,0.0274873742456871);

\path [draw=blue, semithick]
(axis cs:0.5,0.628392081870868)
--(axis cs:0.5,0.628392081870868);

\path [draw=blue, semithick]
(axis cs:1,1.20911167882745)
--(axis cs:1,1.20911167882745);

\path [draw=blue, semithick]
(axis cs:1.5,1.76729396402804)
--(axis cs:1.5,1.76729396402804);

\path [draw=blue, semithick]
(axis cs:2,2.27076417026709)
--(axis cs:2,2.27076417026709);

\path [draw=blue, semithick]
(axis cs:2.5,2.66538498418358)
--(axis cs:2.5,2.66538498418358);

\path [draw=blue, semithick]
(axis cs:3,2.92221664120549)
--(axis cs:3,2.92221664120549);

\addplot [semithick, red, mark=asterisk, mark size=3, mark options={solid}, only marks]
table {%
0 0.0085636
0.5 0.5472
1 1.0785
1.5 1.6038
2 2.1237
2.5 2.6342
3 3.0907
};
\addlegendentry{FEM}
\addplot [semithick, color0, mark=asterisk, mark size=3, mark options={solid}, only marks]
table {%
0 0.0274873742456871
0.5 0.628392081870868
1 1.20911167882745
1.5 1.76729396402804
2 2.27076417026709
2.5 2.66538498418358
3 2.92221664120549
};
\addlegendentry{F/T sensor}
\node[label={RMSE = 0.085 N}] at (axis cs:2.35,-0.2) {};
\end{axis}

\end{tikzpicture}}
				\vspace{-2mm}
		\caption{Shear with triangular flat indenter (3DP)}
	\end{subfigure}
	\vskip \baselineskip
	\vspace{-2mm}
	\begin{subfigure}{0.49\textwidth}
	\scalebox{0.8}{
\begin{tikzpicture}

\definecolor{color0}{rgb}{0.12156862745098,0.466666666666667,0.705882352941177}

\begin{axis}[
legend cell align={left},
legend style={fill opacity=0.8, draw opacity=1, text opacity=1, draw=white!80!black,
at={(0.97,0.87)},},
tick align=outside,
tick pos=left,
x grid style={white!69.0196078431373!black},
xlabel={Horizontal displacement [mm]},
xmin=-0.1, xmax=3.15,
xtick style={color=black},
xtick={-0.5,0,0.5,1,1.5,2,2.5,3,3.5},
xticklabels={\ensuremath{-}0.5,0.0,0.5,1.0,1.5,2.0,2.5,3.0,3.5},
y grid style={white!69.0196078431373!black},
ylabel={Vertical force [N]},
ymin=-4, ymax=0,
ylabel style={yshift=10pt},
ytick style={color=black},
ytick={-4,-3.5,-3,-2.5,-2,-1.5,-1,-0.5,0},
yticklabels={
  \ensuremath{-}4.0,
  \ensuremath{-}3.5,
  \ensuremath{-}3.0,
  \ensuremath{-}2.5,
  \ensuremath{-}2.0,
  \ensuremath{-}1.5,
  \ensuremath{-}1.0,
  \ensuremath{-}0.5,
  0.0
}
]
\path [draw=blue, semithick]
(axis cs:0,-3.67357741872093)
--(axis cs:0,-3.67357741872093);

\path [draw=blue, semithick]
(axis cs:0.5,-3.57943670015473)
--(axis cs:0.5,-3.57943670015473);

\path [draw=blue, semithick]
(axis cs:1,-3.59498991991411)
--(axis cs:1,-3.59498991991411);

\path [draw=blue, semithick]
(axis cs:1.5,-3.53714064181641)
--(axis cs:1.5,-3.53714064181641);

\path [draw=blue, semithick]
(axis cs:2,-3.52752556537268)
--(axis cs:2,-3.52752556537268);

\path [draw=blue, semithick]
(axis cs:2.5,-3.52843263167732)
--(axis cs:2.5,-3.52843263167732);

\path [draw=blue, semithick]
(axis cs:3,-3.50689936402646)
--(axis cs:3,-3.50689936402646);

\addplot [semithick, red, mark=asterisk, mark size=3, mark options={solid}, only marks]
table {%
0 -3.6896
0.5 -3.6869
1 -3.6662
1.5 -3.6361
2 -3.6036
2.5 -3.5785
3 -3.5186
};
\addlegendentry{FEM}
\addplot [semithick, color0, mark=asterisk, mark size=3, mark options={solid}, only marks]
table {%
0 -3.67357741872093
0.5 -3.57943670015473
1 -3.59498991991411
1.5 -3.53714064181641
2 -3.52752556537268
2.5 -3.52843263167732
3 -3.50689936402646
};
\addlegendentry{F/T sensor}
\node[label={RMSE = 0.050 N}] at (axis cs:2.35,-0.56) {};
\end{axis}

\end{tikzpicture}}
			\vspace{-2mm}
	\caption{Shear with triangular flat indenter (3DP)}
	\end{subfigure}
	\begin{subfigure}{0.49\textwidth}
	\vspace{-1mm}
	\scalebox{0.8}{
\begin{tikzpicture}

\definecolor{color0}{rgb}{0.12156862745098,0.466666666666667,0.705882352941177}

\begin{axis}[
legend cell align={left},
legend style={
	fill opacity=0.8,
	draw opacity=1,
	text opacity=1,
	at={(0.023,0.33)},
	anchor=north west,
	draw=white!80!black
},
tick align=outside,
tick pos=left,
x grid style={white!69.0196078431373!black},
xlabel={Indentation depth [mm]},
ylabel style={yshift=15pt},
xmin=0, xmax=2.0875,
xtick style={color=black},
y grid style={white!69.0196078431373!black},
ylabel={Vertical force [N]},
ymin=-1.99625045019666, ymax=0.0303743071522218,
ytick style={color=black},
ytick={-2,-1.75,-1.5,-1.25,-1,-0.75,-0.5,-0.25,0,0.25},
yticklabels={
  \ensuremath{-}2.00,
  \ensuremath{-}1.75,
  \ensuremath{-}1.50,
  \ensuremath{-}1.25,
  \ensuremath{-}1.00,
  \ensuremath{-}0.75,
  \ensuremath{-}0.50,
  \ensuremath{-}0.25,
  0.00,
  0.25
}
]
\path [draw=blue, semithick]
(axis cs:0.25,-0.0726902755896794)
--(axis cs:0.25,-0.0726902755896794);

\path [draw=blue, semithick]
(axis cs:0.5,-0.151783355064188)
--(axis cs:0.5,-0.151783355064188);

\path [draw=blue, semithick]
(axis cs:0.75,-0.29709875265263)
--(axis cs:0.75,-0.29709875265263);

\path [draw=blue, semithick]
(axis cs:1,-0.46883421706539)
--(axis cs:1,-0.46883421706539);

\path [draw=blue, semithick]
(axis cs:1.25,-0.694474493568427)
--(axis cs:1.25,-0.694474493568427);

\path [draw=blue, semithick]
(axis cs:1.5,-1.00978683695961)
--(axis cs:1.5,-1.00978683695961);

\path [draw=blue, semithick]
(axis cs:1.75,-1.40789379959318)
--(axis cs:1.75,-1.40789379959318);

\path [draw=blue, semithick]
(axis cs:2,-1.90413114304444)
--(axis cs:2,-1.90413114304444);

\addplot [semithick, red, mark=asterisk, mark size=3, mark options={solid}, only marks]
table {%
0.25 -0.061745
0.5 -0.15065
0.75 -0.26475
1 -0.41078
1.25 -0.63129
1.5 -0.95072
1.75 -1.358
2 -1.8723
};
\addlegendentry{FEM}
\addplot [semithick, color0, mark=asterisk, mark size=3, mark options={solid}, only marks]
table {%
0.25 -0.0726902755896794
0.5 -0.151783355064188
0.75 -0.29709875265263
1 -0.46883421706539
1.25 -0.694474493568427
1.5 -1.00978683695961
1.75 -1.40789379959318
2 -1.90413114304444
};
\addlegendentry{F/T sensor}
\node[label={RMSE = 0.044 N}] at (axis cs:0.52,-2.01) {};
\end{axis}

\end{tikzpicture}}
			\vspace{-2mm}
	\caption{Pressure with two spherical indenters (SS)}
	\end{subfigure}
	\caption{The plots show the agreement (measured by the root-mean-square error, RMSE) between the total force resulting from FEM indentations (in red) and the readings of an F/T sensor (in blue) when repeating the same indentations in a controlled experimental scenario. While the FEM simulations are performed assuming a fixed friction coefficient, the real-world experiments employed indenters that were either 3D-printed (3DP) or made of stainless steel (SS). In (a), a 3DP spherically-ended indenter was first pressed 2 mm down in the center of the sensing surface. Then it was laterally sheared in one direction, and the horizontal force was recorded at discrete steps. In (b), the same experiment was repeated with an SS indenter with the same geometry. In (c), the shear experiment was carried out with a square flat-ended indenter at a depth of 1 mm, and in (d) with a triangular flat-ended indenter at a depth of 2 mm. The corresponding vertical force for the same experiment as in (d) is shown in (e), where a slight decrease in force was detected during shear from both the FEM simulation and the F/T sensor. In (f), two SS spherically-ended indenters (both attached to the same F/T sensor in the real world) were employed to make vertical indentations on the sensing surface, with the total vertical force recorded at discrete steps. Since the two indenters had a constant difference in height of 1.1 mm, the first four steps (up to 1 mm depth) resulted from contact with only one of the indenters, while the remaining steps (after 1 mm) resulted from a double indentation.}\label{fig:agreement}
\end{figure*}
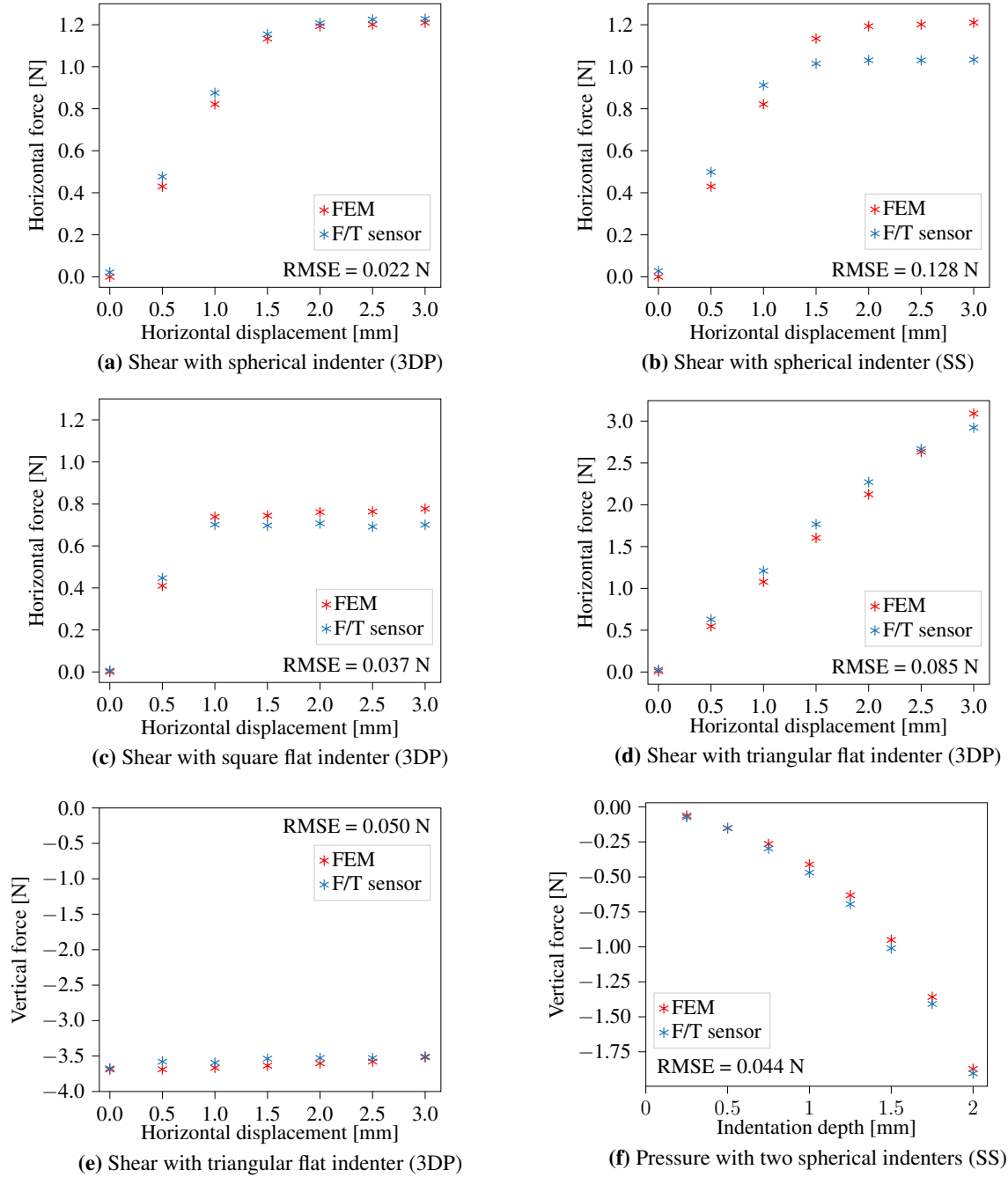

The material characterization followed state-of-the-art techniques based on uniaxial tension, pure shear, and equibiaxial tension tests. The models obtained required no further calibration for the simulations described in this article. In fact, the characterization tests were entirely independent of the evaluation experiments carried out in this work. In order to verify the consistency of the Ogden models and the related FEM simulations, the previous \mbox{work\cite{sferrazza_fem}} also showed an accurate total force agreement when the same vertical indentation experiments were performed both in the FEM environment and the real world, where the total force was obtained from the readings of a commercial six-axis F/T sensor. Such verification experiments were augmented here to test the total force accuracy also in the case of shear-predominant or multi-contact indentations. The results of these experiments are shown in \mbox{Fig.~\ref{fig:agreement},} where the total force resulting from FEM indentations was compared with the force measured by an F/T sensor (ATI Mini27 Titanium with an horizontal resolution of 0.03 N and a vertical resolution of 0.06 N) when repeating the same indentations in the real-world. The real-world experiments were carried out by mounting the F/T sensor and the appropriate indenters to the spindle of a controllable milling machine. The friction coefficient was tuned to a value of 0.9 using a single 3D-printed rough indenter in the experiment shown in (a). However, the remaining experiments showed generalization to different indentation shapes (see (b)) and a limited loss of accuracy for indenters of a different and smoother material, such as stainless steel (see (c)). In addition, the friction coefficient has a limited influence on the accuracy of the vertical component of the total force, as shown for the multi-contact experiment performed in (d) with stainless steel indenters.

The indentation trajectories collected in the FEM environment, which were  employed to generate training data, were randomized as follows:
\begin{enumerate}
	\item First, for each trajectory, one of the 21 indenters modeled in simulation was selected randomly and translated to a random horizontal position over the sensing surface with a randomized orientation.
	\item Then, 80\% of the time, a vertical indentation followed by randomized horizontal translations was simulated. In the remaining 20\% of the time, random 3D displacements were directly simulated in the vicinity of the initial position. Such a split in the training trajectories aimed to favor the typical robotic manipulation case, where shear motion happens after a vertical grasp. Each indentation trajectory was split into 50 steps, with the maximum intra-step displacement constrained to 0.1 mm to facilitate convergence. The maximum depth reached by the indenters was 2 mm, while the maximum lateral displacement from the start of the indentation was 3 mm. Static steps were employed, therefore neglecting time-dependent material effects, which are however limited, as shown in \mbox{Fig.~\ref{fig:force_time}.}
\end{enumerate}

\section{Projection of a particle onto the image plane}
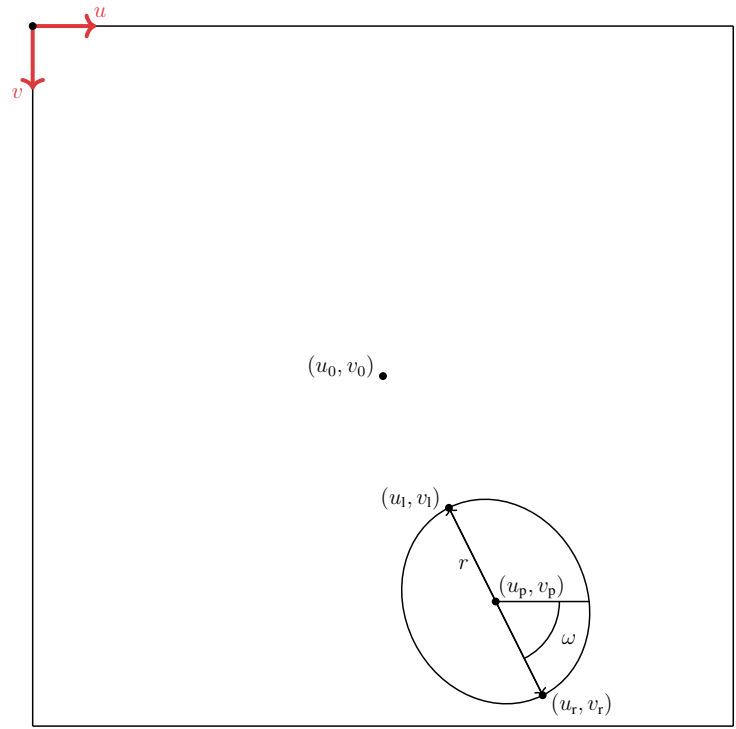
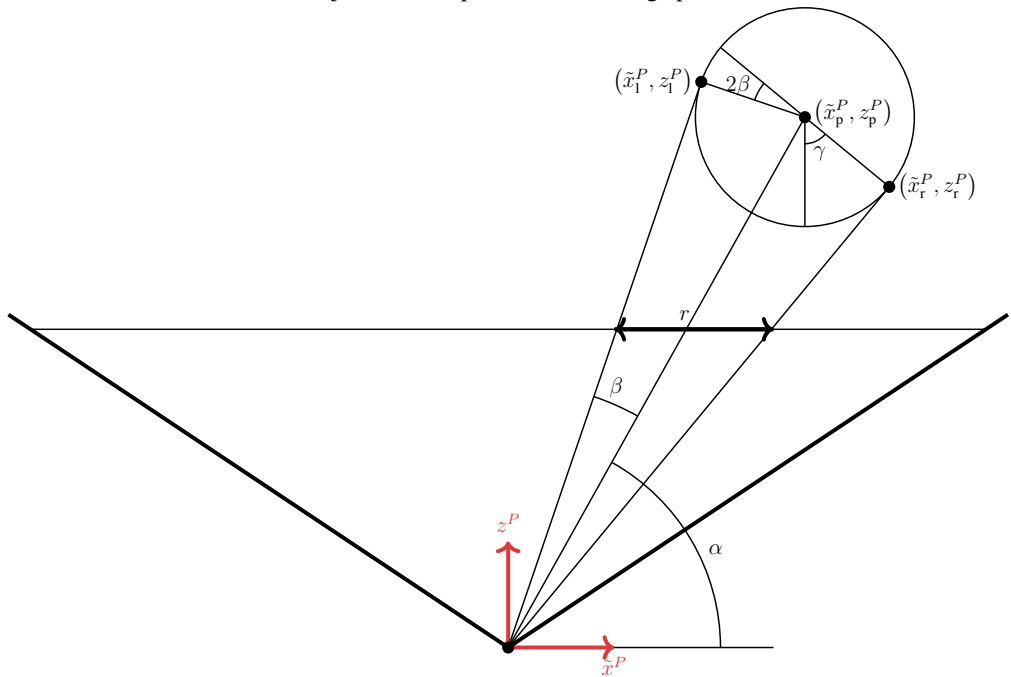
\begin{figure*}
	\centering
	\begin{subfigure}{\textwidth}
		\centering
	\scalebox{0.75}{\definecolor{red}{RGB}{220,57,61}

\def\R{6.59}  
\def\r{5.72}
\def\omeg{63.4359}

\def\centerarc[#1](#2)(#3:#4:#5)[#6](#7)
{ \draw[#1] ($(#2)+({#5*cos(#3)},{#5*sin(#3)})$) arc (#3:#4:#5) node[#6]{#7}; }

\begin{tikzpicture}[scale=0.3, axis/.style={->,dashed},thick]

\node[coordinate] (c) at (29.08,-36.17){};
\node[coordinate] (r) at (32.03,-42.06){};
\node[coordinate] (l) at (26.14,-30.28){};
	
\draw[->] (l) -- (r) node[midway,left,yshift=20,xshift=-10]{$r$};
\draw[->] (r) -- (l);
\draw (c) -- (35,-36.17);

\draw (0,0) -- (44,0);
\draw (0,0) -- (0,-44);
\draw (44,0) -- (44,-44);
\draw (0,-44) -- (44,-44);

\draw[->,red,line width=0.75mm] (0,0) -- (4,0) node[above,xshift=2]{$u$};
\draw[->,red,line width=0.75mm] (0,0) -- (0,-4) node[left,yshift=-2]{$v$};

\centerarc[](c)(0:-\omeg:4)[right,midway,xshift=2,yshift=-2]($\omega$);

\draw (c) circle [x radius=\R, y radius=\r, rotate=-\omeg];

\filldraw (0,0) circle (0.2);
\filldraw (c) circle (0.2) node[right,yshift=8,xshift=-3]{$\left(u_\text{p},v_\text{p}\right)$};
\filldraw (r) circle (0.2) node[right,yshift=-5]{$\left(u_\text{r},v_\text{r}\right)$};
\filldraw (l) circle (0.2) node[left,yshift=5]{$\left(u_\text{l},v_\text{l}\right)$};
\filldraw (22,-22) circle (0.2) node[left,yshift = 5]{$\left(u_0,v_0\right)$};

\end{tikzpicture}}	
	\caption{Projection of a sphere onto the image plane}	
	\end{subfigure}
	\begin{subfigure}{\textwidth}
				\centering
	\scalebox{0.75}{\definecolor{red}{RGB}{220,57,61}

\def\R{4.125}  
\def\alph{60.7941}
\def\bet{10.3715}

\def\centerarc[#1](#2)(#3:#4:#5)[#6](#7)
{ \draw[#1] ($(#2)+({#5*cos(#3)},{#5*sin(#3)})$) arc (#3:#4:#5) node[#6]{#7}; }

\begin{tikzpicture}[scale=0.5, axis/.style={->,dashed},thick]

\node[coordinate] (c) at (11.1803,20){};
\node[coordinate] (r) at (14.3597, 17.3719){};
\node[coordinate] (rt) at (8.0009, 22.6281){};
\node[coordinate] (l) at (7.2762, 21.3317){};
	
\draw (0,0) -- (10,0);
\draw (c) -- (11.1803,20-4.125);
\draw (c) -- (rt);

\draw[->,red,line width=0.75mm] (0,0) -- (4,0) node[below]{$\tilde{x}^P$};
\draw[->,red,line width=0.75mm] (0,0) -- (0,4) node[above]{$z^P$};

\draw (-18,12) -- (18,12);
\draw [line width=0.75mm] (0,0) -- (18.8321,12.5547);
\draw [line width=0.75mm] (0,0) -- (-18.8321,12.5547);
\draw [->, line width=0.75mm] (4,12) -- (10,12);
\draw [->, line width=0.75mm] (10,12) -- (4,12) node[above,midway,xshift=-5]{$r$};
	
\draw (c) circle (\R);

\draw (0,0) -- (r);
\draw (0,0) -- (l);

\draw (c) -- (r);
\draw (c) -- (l);

\centerarc[](0,0)(0:\alph:8)[right,midway,xshift=5,yshift=-5]($\alpha$);
\centerarc[](0,0)(\alph:\alph+\bet:10)[midway,above]($\beta$);
\centerarc[](c)(270:270+\alph-\bet:1)[xshift=-3,yshift=-2,below]($\gamma$);
\centerarc[](c)(90+\alph-\bet:90+\alph+\bet:2)[midway,yshift=3,left]($2\beta$);

\filldraw (0,0) circle (0.2);
\filldraw (c) circle (0.2) node[right,yshift=2]{$\left(\tilde{x}_\text{p}^P,z_\text{p}^P\right)$};
\filldraw (r) circle (0.2) node[right]{$\left(\tilde{x}_\text{r}^P,z_\text{r}^P\right)$};
\filldraw (l) circle (0.2) node[left]{$\left(\tilde{x}_\text{l}^P,z_\text{l}^P\right)$};

\draw (c) -- (0, 0);
\end{tikzpicture}}	
	\caption{2D view of the projection of a sphere}
	\end{subfigure}
	\caption{The figures show that the projection of a sphere corresponds to an ellipse (see (a)) in the image plane. The length $r$ of the major axis of this ellipse can be computed via 2D geometry in the plane that contains the optical axis and the ray passing through the camera and the center of the sphere.}
	\label{fig:ellipse_projection}
\end{figure*}

The projection of the spherical particle centered at \mbox{$s_\text{p}^P := \left(x_\text{p}^P,y_\text{p}^P,z_\text{p}^P\right)$} via the pinhole camera model results in an ellipse on the image plane\cite{sphere_projection}, see Fig.~\ref{fig:ellipse_projection}(a). 
The pixel length $r$ of the major axis of each ellipse can be computed via the projection formulas in the plane containing the camera's optical axis and the camera ray passing through the  center of the spherical particle. An example projection in this plane is shown in Fig.~\ref{fig:ellipse_projection}(b). The coordinate $\tilde{x}_\text{p}^P$ can be computed from the horizontal coordinates of the center of the particle as,
\begin{gather}
\label{eq:coord_transformation}
\tilde{x}_\text{p}^P = \sqrt{\left(x_\text{p}^P\right)^2+\left(y_\text{p}^P\right)^2}.
\end{gather}
Then, from the figure, it follows that:
\begin{gather}
\alpha = \arctan\left(\dfrac{z_\text{p}^P}{\tilde{x}_\text{p}^P}\right), \\
\beta = \arcsin\left(\dfrac{R}{\sqrt{\left(\tilde{x}_\text{p}^P\right)^2+\left(z_\text{p}^P\right)^2}}\right), \\
\gamma = \alpha - \beta,
\end{gather}
where $R$ is the radius of the sphere.
The pixel length $r$ of the major axis can then be computed as:
\begin{gather}
\tilde{x}_\text{r}^P = \tilde{x}_\text{p}^P+R\sin\gamma, \\
z_\text{r}^P = z_\text{p}^P-R\cos\gamma ,\\
\tilde{x}_\text{l}^P = \tilde{x}_\text{p}^P-R\sin(\gamma+2\beta), \\
z_\text{l}^P = z_\text{p}^P+R\cos(\gamma+2\beta) ,\\\label{eq:major_axis}
r = \left|f\left(\dfrac{\tilde{x}_\text{r}^P}{z_\text{r}^P} - \dfrac{\tilde{x}_\text{l}^P}{z_\text{l}^P} \right)\right|.
\end{gather}
%

The orientation of the ellipse on the image plane is fully determined by the horizontal position of the particle, and can therefore be computed trivially as:
\begin{gather}
\omega = \arctantwo\left(y_\text{p}^P,x_\text{p}^P\right).
\end{gather}
Additionally, the center of the ellipse can be computed by observing that $(u_\text{r},v_\text{r})$ and $(u_\text{l},v_\text{l})$ correspond to the projection of $(\tilde{x}_\text{r}^P,z_\text{r}^P)$ and $(\tilde{x}_\text{l}^P,z_\text{l}^P)$, respectively, onto the image plane:
\begin{gather}
x_\text{r}^P = \tilde{x}_\text{r}^P  \cos \omega, \quad
y_\text{r}^P = \tilde{x}_\text{r}^P  \sin \omega, \\
u_\text{r} = f \dfrac{x_\text{r}^P}{z_\text{r}^P} + u_0, \quad
v_\text{r} = f \dfrac{y_\text{r}^P}{z_\text{r}^P} + v_0, \\
x_\text{l}^P = \tilde{x}_\text{l}^P  \cos \omega, \quad
y_\text{l}^P = \tilde{x}_\text{l}^P  \sin \omega, \\
u_\text{l} = f \dfrac{x_\text{l}^P}{z_\text{l}^P} + u_0, \quad
v_\text{l} = f \dfrac{y_\text{l}^P}{z_\text{l}^P} + v_0,
\end{gather}
where $(u_0,v_0)$ are the coordinates of the pinhole image center.

Therefore, the pixel coordinates of the center of the ellipse are:
\begin{gather}
u_\text{p} = \dfrac{u_\text{r}+u_\text{l}}{2}, \\
v_\text{p} = \dfrac{v_\text{r}+v_\text{l}}{2}.
\end{gather}
Finally, noting that the pixel length of the minor axis of the ellipses does not vary with the horizontal coordinates of the sphere\cite{sphere_projection}, this length can be computed for a trivial case, that is, when the center of a particle lies on the optical axis (i.e., $x_\text{p}^P = y_\text{p}^P = 0$). The same formulas as in \eqref{eq:coord_transformation}-\eqref{eq:major_axis} can be employed, since for this special case the projection results in a circle, where both the major axis and the minor axis of the ellipse correspond to the diameter. Using the center, the axis lengths and the orientation of each ellipse, these can be drawn using the drawing functionality of OpenCV\footnote{https://opencv.org/}. 

\section{Remapping}
		
As shown in Fig.~\ref{fig:frames_definition}, for a pixel $p := (u,v)$ in the image plane of the pinhole camera, a 3D point $s^P := (x^P,y^P,t_z^{GP})$ was retrieved using the pinhole projection equations as:
\begin{gather}
x^P = \dfrac{t_z^{GP}}{f}(u-u_0), \\
y^P = \dfrac{t_z^{GP}}{f}(v-v_0).
\end{gather}
The 3D point was then converted to the coordinate system of the real-world camera, indicated with the superscript $C$, through the corresponding rotation and translation operations:
\begin{align}
s^C = R^{GC}\left(R^{GP}\right)^{-1}\left(s^P - t^{GP}\right) + t^{GC}.
\end{align}
The corresponding pixel in the real-world image was then retrieved via the transformation function obtained from the calibration toolbox.
\vspace{8mm}
\begin{center}
	\scalebox{0.8}{\def\r{8}
\definecolor{green}{RGB}{54,181,80}
\definecolor{red}{RGB}{220,57,61}
\begin{tikzpicture}[scale=0.15,thick,x=1mm,y=1mm]

	\filldraw[green] (278, 496) circle (\r);
	\filldraw[green] (276.808, 524.2324) circle (\r);
	\filldraw[green] (278.2292, 465.2516) circle (\r);
	\filldraw[green] (362.7921, 464.541) circle (\r);
	\filldraw[green] (318.7341, 463.8304) circle (\r);
	\filldraw[green] (383.3998, 499.361) circle (\r);
	\filldraw[green] (439.0365, 499.2034) circle (\r);
	\filldraw[green] (445.9337, 459.5667) circle (\r);
	\filldraw[green] (480.7537, 467.3834) circle (\r);
	\filldraw[green] (509.1782, 465.9622) circle (\r);
	\filldraw[green] (542.5769, 468.0941) circle (\r);
	\filldraw[green] (477.9112, 497.2292) circle (\r);
	\filldraw[green] (533.339, 526.3643) circle (\r);
	\filldraw[green] (548.9724, 491.5443) circle (\r);
	\filldraw[green] (577.3969, 520.6794) circle (\r);
	\filldraw[green] (583.0818, 464.541) circle (\r);
	\filldraw[green] (623.5867, 494.3867) circle (\r);
	\filldraw[green] (651.3006, 498.6504) circle (\r);
	\filldraw[green] (655.5643, 526.3643) circle (\r);
	\filldraw[green] (689.6737, 525.6536) circle (\r);
	\filldraw[green] (666.2235, 472.3577) circle (\r);
	\filldraw[green] (687.5418, 459.5667) circle (\r);
	\filldraw[green] (419.641, 531.3385) circle (\r);
	\filldraw[green] (467.9627, 527.7855) circle (\r);
	\filldraw[green] (411.1137, 465.2516) circle (\r);
	\filldraw[green] (314.4704, 525.6536) circle (\r);
	\filldraw[green] (308.0749, 490.123) circle (\r);
	\filldraw[green] (384, 528) circle (\r);
	\filldraw[green] (512, 512) circle (\r);
	\filldraw[green] (592, 496) circle (\r);
	\filldraw[green] (608, 528) circle (\r);
	\filldraw[green] (624, 464) circle (\r);
	\filldraw[green] (352, 512) circle (\r);
	\filldraw[green] (352, 512) circle (\r);

    \filldraw (480, 192) circle (\r);
    \filldraw (672, 160) circle (\r);
    \filldraw (663.055, 240.5046) circle (\r);	
    \filldraw (530.1421, 272.2273) circle (\r);
	
	\draw[line width = 2] (192, 128) -- (192, 448);
	\draw[line width = 2] (192, 448) -- (256, 448);
	\draw[line width = 2] (256, 448) -- (256, 544) 
	-- (704, 544) -- (704, 448);
	\draw[line width = 2] (704, 448) -- (768, 448);
	\draw[line width = 2] (768, 448) -- (768, 128);
	\draw[dashed] (256, 448) -- (704, 448);
	\draw[line width = 1.5] (480, 192) -- (352, 288);
	\draw[line width = 1.5] (480, 192) -- (608, 288);
	\draw (372.9707, 272.2719) -- (586.6296, 271.9722)
	-- (586.857, 272.1427);


	\draw[line width = 1.5] (672, 160) -- (736, 272);
	\draw[line width = 1.5] (672, 160) -- (544, 224);
	\draw (726.9943, 256.24) -- (561.9927, 215.0036);
	\draw[shift={(671.891, 159.941)}, rotate=18.4441, red, ->,line width = 1.2] (0, 0) -- node [left,pos=1.1,xshift=5]{$z^C$}(0, 48);
	\draw[shift={(671.891, 159.941)}, rotate=18.4441, red, ->,line width = 1.2] (0, 0) -- node [right,pos=0.9]{$x^C$} (48, 0);
	\draw[dotted] (672, 160) -- (640, 448);
	\draw[dotted] (640, 448) -- (480, 192);
	\draw[->] (480, 192) -- node [left,pos=0.8]{ $t_z^{GP}$} (480, 448);

	\draw[red, ->, line width = 1.2] (256, 448) -- node [left,pos=1] {$z^G$} (256, 496);
	\draw[red, ->, line width = 1.2] (256, 448) -- node [below,pos=0.8]{$x^G$} (304, 448);
	\draw[red, ->, line width = 1.2] (480, 192) -- node [left,xshift=4,pos=1.2]{$z^P$} (480, 240);
	\draw[red, ->, line width = 1.2] (480, 192) -- node [right,pos=0.8,yshift=-2]{$x^P$} (528, 192);
	
\end{tikzpicture}}
	
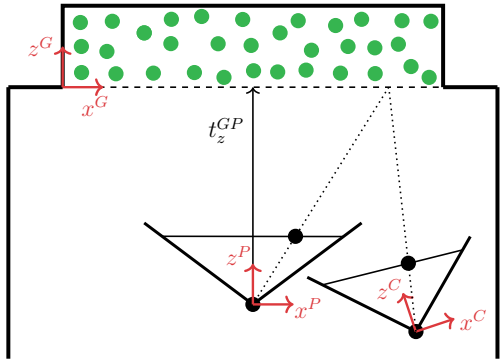
\captionof{figure}{In the figure, a pixel in the pinhole camera is mapped to the corresponding pixel in the real-world camera.}\label{fig:frames_definition}
\end{center}	

\section{Results}
\begin{figure*}
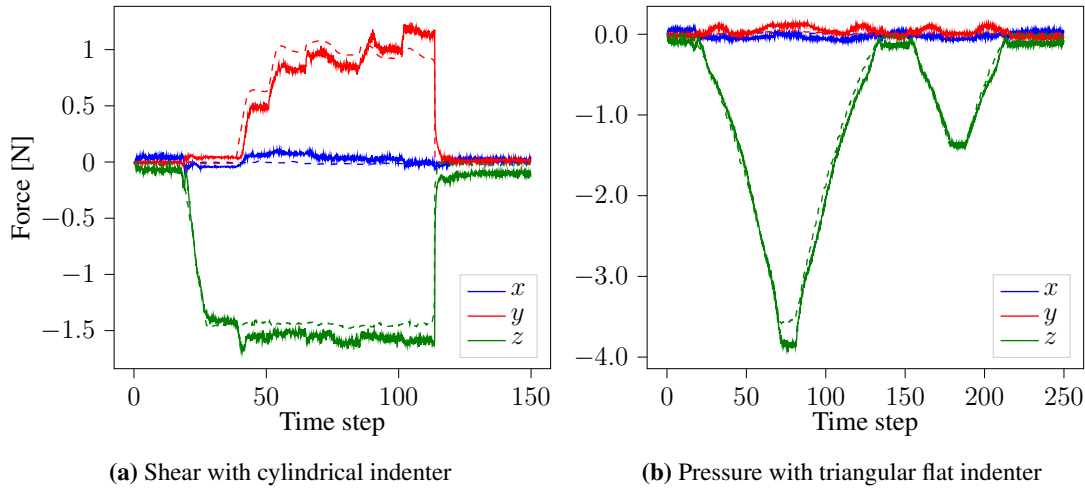

	\centering
	\begin{subfigure}{0.45\textwidth}
		\scalebox{0.9}{\input{img/shear_time_ind4.tex}}	
		\caption{Shear with cylindrical indenter}
	\end{subfigure}
	\begin{subfigure}{0.45\textwidth}
		\scalebox{0.9}{\input{img/pressure_time_ind2.tex}}	
		\caption{Pressure with triangular flat indenter}
	\end{subfigure}
	\caption{The plots compare the total force computed from the predictions of the neural network (solid lines) against the readings of an F/T sensor (dashed lines) for each of the three force components. In (a), a thin cylindrical indenter was first pressed against the sensing surface up to a depth of 1 mm, then it was laterally displaced in the $y$ direction in discrete steps up to 3 mm, and finally lifted. In (b), two pressure cycles were executed with a triangular flat indenter, first up to 2 mm, then up to 1 mm. Although the network was only trained on static data, the predictions accurately capture the force trends in both the figures. The main inaccuracies can be observed for considerably larger deformations (where the material characterization in previous work\cite{sferrazza_fem} showed larger variance) with the triangular flat indenter, or during the unloading phase (where the material showed mild relaxation effects). }\label{fig:force_time}
\end{figure*}

\begin{figure}[H]
	\vspace{-1mm}
	\centering
	\begin{subfigure}{0.45\columnwidth}
		\includegraphics[width=0.9\linewidth]{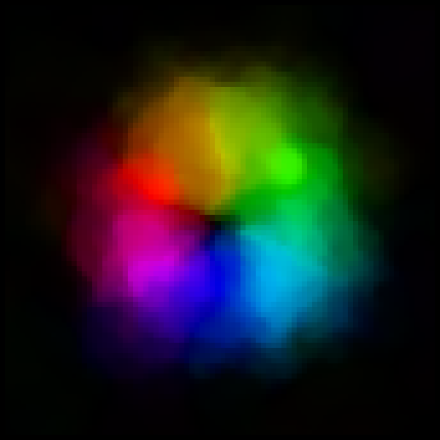}	
		\caption{Simulation}
	\end{subfigure}
	\begin{subfigure}{0.45\columnwidth}
		\hspace{1cm}
		\includegraphics[width=0.9\linewidth]{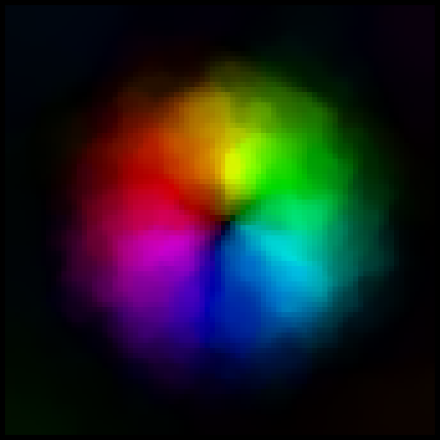}
		\caption{Reality}
	\end{subfigure}
	\vspace{-3mm}
	\caption{Comparison of the optical flow obtained in simulation (a) versus reality (b). The color represents the direction, while darker regions represent smaller displacements.}\label{fig:optical_flow}
	\vspace{-7mm}
\end{figure}

This section presents supplementary results and illustrations in addition to those in the main article. \mbox{Fig.~\ref{fig:optical_flow}} compares optical flow examples obtained for the same indentation in simulation and reality. In \mbox{Fig.~\ref{fig:force_time}}, a programmable milling machine was employed to make indentations using two of the test indenters, and the total force recorded with an F/T sensor was compared with the real-time prediction of the neural network presented in the main article. 

Table \mbox{\ref{table:metrics_indentation}} and Table \mbox{\ref{table:metrics_indenter}} show in detail the different error metrics listed by data subgroups for the real-world test dataset, depending on the type of indentation or the indenter employed. Both tables are based on metrics computed using the raw-feature model described in the main article. It can be noted that shear-dominant indentations may present a decrease in accuracy in the prediction of the $z$ component of the force. This is partly due to the fact that shear-dominant data generally showed higher noise, as a small misalignment in the indenter mounting may lead to considerably different behavior of the material during the shearing trajectory. In addition, as shown in the main article, the model tended to generalize to multi-contact indentations. However, the performance for such contact conditions may be further improved by including a portion of multi-contact data in the training dataset. Among the six indenters employed, the results generally showed a correlation between the range of forces and the errors recorded. In addition, it turned out to be very challenging to accurately align the tilted-plane indenter (which is the indenter that shows a deeper side in Fig. 2 of the main article) with the reference system of the gel for data collection. For this reason, shear data were not collected with such an indenter.

\begin{table*}[]
	\centering
	\begin{tabular}{llccc}
		\hline
		&                          & \multicolumn{1}{c}{Vertical} & \multicolumn{1}{c}{Shear-dominant} & \multicolumn{1}{c}{Multi-contact} \\ \hline
		\multicolumn{1}{l|}{\multirow{3}{*}{RMSE}}      & \multicolumn{1}{l|}{$x$} &          0.003                        &              0.010                     &               0.004                    \\ \cline{2-5} 
		\multicolumn{1}{l|}{}                               & \multicolumn{1}{l|}{$y$} &   0.005                           &          0.011                          &          0.004                         \\ \cline{2-5} 
		\multicolumn{1}{l|}{}                               & \multicolumn{1}{l|}{$z$} &      0.013                        &       0.018                             &         0.013                          \\ \hline
		\multicolumn{1}{l|}{\multirow{3}{*}{RMSET}}      & \multicolumn{1}{l|}{$x$} &        0.034                         &        0.548                            &        0.088                           \\ \cline{2-5} 
		\multicolumn{1}{l|}{}                               & \multicolumn{1}{l|}{$y$} &       0.057                       &        0.606                           &        0.061                           \\ \cline{2-5} 
		\multicolumn{1}{l|}{}                               & \multicolumn{1}{l|}{$z$} &       0.277                       &         0.465                           &         0.488                          \\ \hline
		\multicolumn{1}{l|}{\multirow{3}{*}{MAE (bin)}}     & \multicolumn{1}{l|}{$x$} &     0.001                         &        0.002                            &      0.001                             \\ \cline{2-5} 
		\multicolumn{1}{l|}{}                               & \multicolumn{1}{l|}{$y$} &     0.001                         &       0.002                             &     0.001                              \\ \cline{2-5} 
		\multicolumn{1}{l|}{}                               & \multicolumn{1}{l|}{$z$} &     0.002                         &        0.004                            &     0.002                              \\ \hline
		\multicolumn{1}{l|}{\multirow{3}{*}{MAE (total)}}   & \multicolumn{1}{l|}{$x$} &          0.022                    &        0.333                            &     0.060                              \\ \cline{2-5} 
		\multicolumn{1}{l|}{}                               & \multicolumn{1}{l|}{$y$} &       0.035                       &     0.342                               &      0.046                             \\ \cline{2-5} 
		\multicolumn{1}{l|}{}                               & \multicolumn{1}{l|}{$z$} &       0.168                       &        0.368                            &       0.345                            \\ \hline
		\multicolumn{1}{l|}{\multirow{3}{*}{SDAE (bin)}}    & \multicolumn{1}{l|}{$x$} &     0.003                         &       0.010                             &      0.004                             \\ \cline{2-5} 
		\multicolumn{1}{l|}{}                               & \multicolumn{1}{l|}{$y$} &     0.004                         &      0.011                              &      0.004                             \\ \cline{2-5} 
		\multicolumn{1}{l|}{}                               & \multicolumn{1}{l|}{$z$} &      0.012                        &          0.017                          &       0.012                            \\ \hline
		\multicolumn{1}{l|}{\multirow{3}{*}{SDAE (total)}}  & \multicolumn{1}{l|}{$x$} &        0.025                      &       0.436                             &   0.065                                \\ \cline{2-5} 
		\multicolumn{1}{l|}{}                               & \multicolumn{1}{l|}{$y$} &       0.045                       &        0.500                            &        0.040                           \\ \cline{2-5} 
		\multicolumn{1}{l|}{}                               & \multicolumn{1}{l|}{$z$} &       0.221                       &        0.284                            &      0.345                             \\ \hline
		\multicolumn{1}{l|}{\multirow{3}{*}{Range (total)}} & \multicolumn{1}{l|}{$x$} &       -0.2--0.2                       &       -3.2--3.2                             &       -0.3--0.3                            \\ \cline{2-5} 
		\multicolumn{1}{l|}{}                               & \multicolumn{1}{l|}{$y$} &       -0.6--0.6                       &      -3.8--3.8                              &      -0.3--0.3                             \\ \cline{2-5} 
		\multicolumn{1}{l|}{}                               & \multicolumn{1}{l|}{$z$} &        -4.5--0                      &       -3.8--0                             &      -3.5--0                             \\ \hline
	\end{tabular}
	\caption{The table reports in detail the different error metrics (using a raw-feature model) for each of the three types of real-world indentations, that is, vertical, shear-dominant, and multi-contact indentations. The abbreviations are defined as in the main article. The Newton unit was omitted here for all the values.}
	\label{table:metrics_indentation}
\end{table*}

\begin{table*}
	\vspace{-0.5cm}
	\begin{tabular}{llcccccc}
		\hline
		&                          & \multicolumn{1}{c}{Spherical large} & \multicolumn{1}{c}{Triangular} & \multicolumn{1}{c}{Square} & \multicolumn{1}{c}{Cylindrical} & \multicolumn{1}{c}{Spherical small} & \multicolumn{1}{c}{Tilted-plane} \\ \hline
		\multicolumn{1}{l|}{\multirow{3}{*}{RMSE}}          & \multicolumn{1}{l|}{$x$} &    0.003                                 &         0.009                        &      0.007                       &      0.007                           &      0.004                               &      0.002                            \\ \cline{2-8} 
		\multicolumn{1}{l|}{}                               & \multicolumn{1}{l|}{$y$} &    0.003                                 &          0.008                      &    0.007                         &         0.014                        &  0.003                                   &  0.002                                \\ \cline{2-8} 
		\multicolumn{1}{l|}{}                               & \multicolumn{1}{l|}{$z$} &    0.008                                 &        0.022                          &       0.015                     &    0.021                             &       0.008                              &             0.014                     \\ \hline
		\multicolumn{1}{l|}{\multirow{3}{*}{RMSET}}         & \multicolumn{1}{l|}{$x$} &   0.194                                 &       0.411                         &    0.306                        &        0.494                         &         0.082                            &       0.034                           \\ \cline{2-8} 
		\multicolumn{1}{l|}{}                               & \multicolumn{1}{l|}{$y$} &  0.129                                   &    0.449                            &   0.293                         &          0.644                       &  0.027                                  &      0.047                            \\ \cline{2-8} 
		\multicolumn{1}{l|}{}                               & \multicolumn{1}{l|}{$z$} &    0.154                               &      0.478                          &       0.231                     &      0.498                           &      0.082                               &     0.664                             \\ \hline
		\multicolumn{1}{l|}{\multirow{3}{*}{MAE (bin)}}     & \multicolumn{1}{l|}{$x$} &   0.001                                 &     0.002                           &   0.001                         &     0.001                            &    0.001                                 &     0.001                             \\ \cline{2-8} 
		\multicolumn{1}{l|}{}                               & \multicolumn{1}{l|}{$y$} &    0.001                               &          0.002                      &                           0.001 &                                0.003 &    0.001                                 &                            0.001      \\ \cline{2-8} 
		\multicolumn{1}{l|}{}                               & \multicolumn{1}{l|}{$z$} &    0.001                               &        0.005                        &                           0.003 &             0.005                    &    0.001                                 &                            0.003      \\ \hline
		\multicolumn{1}{l|}{\multirow{3}{*}{SDAE (bin)}}    & \multicolumn{1}{l|}{$x$} &   0.004                                  &         0.009                       &     0.007                       &      0.007                           &          0.003                           &                    0.002              \\ \cline{2-8} 
		\multicolumn{1}{l|}{}                               & \multicolumn{1}{l|}{$y$} &        0.003                             &              0.008                  &                           0.007 &                                0.014 &               0.003                      &  0.002                                \\ \cline{2-8} 
		\multicolumn{1}{l|}{}                               & \multicolumn{1}{l|}{$z$} &             0.007                        &      0.021                          &                           0.014 &                                0.020 &    0.008                                 &                            0.013      \\ \hline
		\multicolumn{1}{l|}{\multirow{3}{*}{MAE (total)}}   & \multicolumn{1}{l|}{$x$} &     0.080                                &      0.197                          &     0.128                       &      0.219                           &        0.030                             &       0.023                           \\ \cline{2-8} 
		\multicolumn{1}{l|}{}                               & \multicolumn{1}{l|}{$y$} &   0.057                                  &        0.208                        &    0.142                        &       0.336                          &        0.021                             &     0.037                             \\ \cline{2-8} 
		\multicolumn{1}{l|}{}                               & \multicolumn{1}{l|}{$z$} &  0.111                                   &     0.379                           &      0.185                      &       0.395                          &       0.058                              &    0.596                              \\ \hline
		\multicolumn{1}{l|}{\multirow{3}{*}{SDAE (total)}}  & \multicolumn{1}{l|}{$x$} &   0.177                                  &   0.361                            &                           0.278 &            0.443                     &       0.076                              &       0.025                           \\ \cline{2-8} 
		\multicolumn{1}{l|}{}                               & \multicolumn{1}{l|}{$y$} &   0.116                                  &     0.398                           &   0.257                         &   0.550                              &      0.017                               &   0.030                               \\ \cline{2-8} 
		\multicolumn{1}{l|}{}                               & \multicolumn{1}{l|}{$z$} &    0.106                                 &   0.291                             &       0.138                     &      0.302                           &     0.059                                &    0.293                              \\ \hline
		\multicolumn{1}{l|}{\multirow{3}{*}{Range (total)}} & \multicolumn{1}{l|}{$x$} &  -1.2--1.2                                   &      -3.2--3.2                          &     -2.2--2.2                       &  -3.2--3.2                               &   -0.1--0.1                                  &   -0.01--0.01                               \\ \cline{2-8} 
		\multicolumn{1}{l|}{}                               & \multicolumn{1}{l|}{$y$} & -1.2--1.2                                    &   -3.2--3.2                             &   -2.2--2.2                         &    -3.8--3.8                             & -0.1--0.1                                    &    -0.01--0.01                              \\ \cline{2-8} 
		\multicolumn{1}{l|}{}                               & \multicolumn{1}{l|}{$z$} &     -1.7--0                                &   -4.3--0                             &   -2.0--0                         &   -4.6--0                              &     -1.0--0                                &  -1.3--0                                \\ \hline
	\end{tabular}
	\caption{The table reports in detail the different error metrics (using a raw-feature model) for each of the six real-world indenters employed to collect the test dataset (and shown in the main article). The Newton unit was omitted here for all the values.}
	\vspace{-0.5cm}
	\label{table:metrics_indenter}
\end{table*}

\begin{figure*}
\begin{tikzpicture}
\setlength{\fboxsep}{0.15cm}%
\setlength{\fboxrule}{0pt}%
\node[] (image) at (0,0){\fbox{\includegraphics[width=0.95\textwidth]{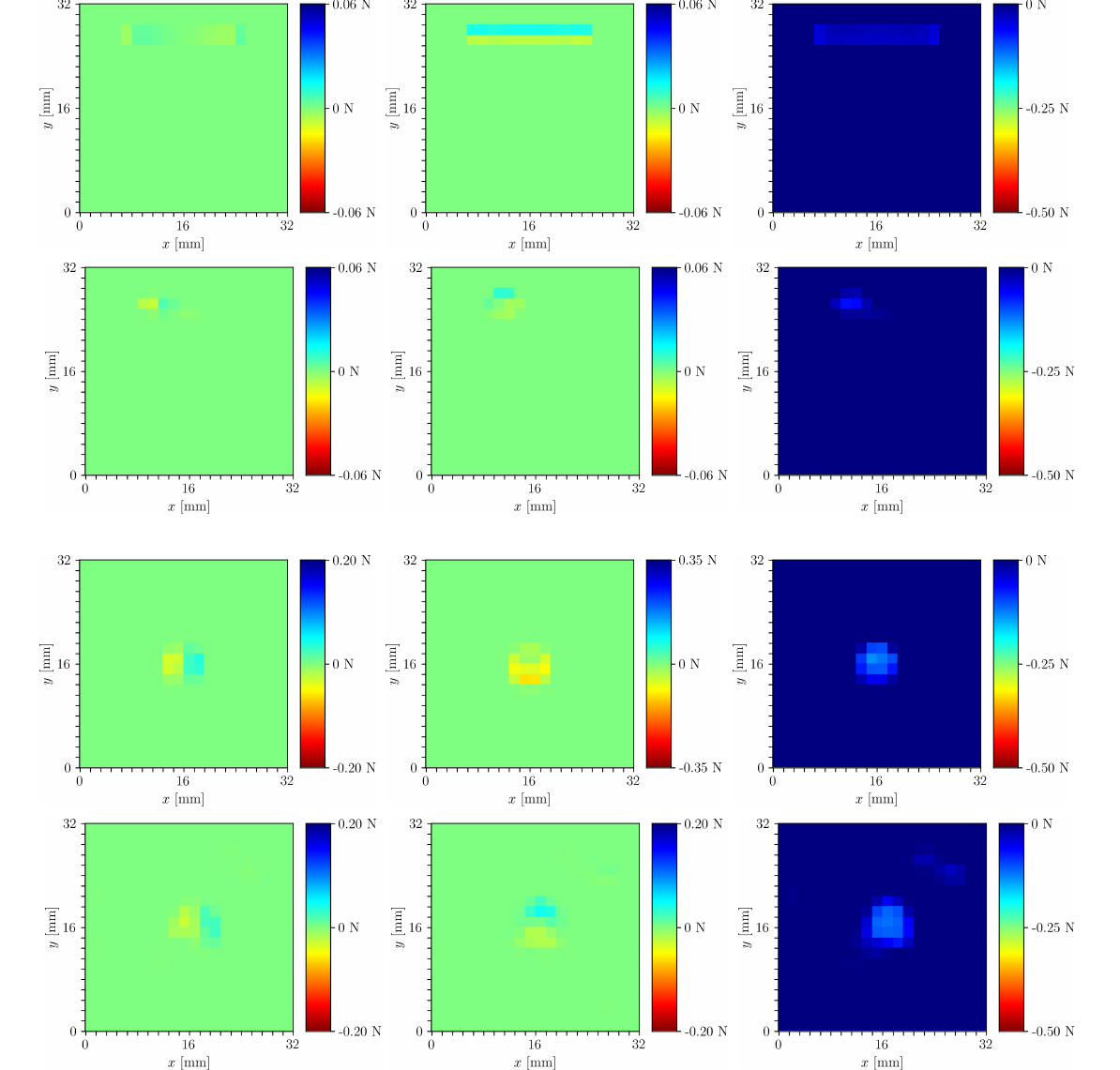}}};
\node[label={[rotate=90]\footnotesize Sample \#1}] (label1) at (-7.8,4.2){};
\node[label={[rotate=-90]\footnotesize Prediction\cite{sim2real_iros}}] (label1p) at (7.6,2.3){};
\node[label={[rotate=-90]\footnotesize Ground Truth}] (label1g) at (7.6,6.2){};
\node[label={[rotate=90]\footnotesize Sample \#2}] (label2) at (-7.8,-4.1){};
\node[label={[rotate=-90]\footnotesize Prediction\cite{sim2real_iros}}] (label2p) at (7.6,-6.){};
\node[label={[rotate=-90]\footnotesize Ground Truth}] (label2g) at (7.6,-2.1){};
\node[label={[]\footnotesize $x$ component}] (labelx) at (-5.3,-8.6){};
\node[label={[]\footnotesize $y$ component}] (labely) at (-0.25,-8.6){};
\node[label={[]\footnotesize $z$ component}] (labelz) at (4.8,-8.6){};
\end{tikzpicture}
	\caption{The figures show the ground truth (first and third rows) and predicted (second and fourth rows) force distribution components ($x$ in the first column, $y$ in the second column, and $z$ in the third column) for the first two samples shown in Fig. 8 in the main article, collected with two different indenters in the real world. Predictions were made with the model trained in previous work\mbox{\cite{sim2real_iros}}, where only vertical indentations made with a spherically-ended indenter were contained in the training dataset. The first two rows show a vertical indentation with a cylindrical indenter; the third and fourth rows show a shear-dominant indentation with a spherically-ended indenter. Note how the model trained in previous work\mbox{\cite{sim2real_iros}} does not generalize well to such cases.}\label{fig:failure_cases}
\end{figure*}

Furthermore, \mbox{Fig.~\ref{fig:failure_cases}} shows that a diverse dataset is crucial for generalization. The samples in the figure correspond to the first two in Fig.~8 of the main article, but the predictions were made with the network trained in previous work\mbox{\cite{sim2real_iros}.} This network was only trained with vertical indentations made with a spherically-ended indenter. To evaluate the generalization, the first two rows show a vertical indentation made with a cylindrical indenter, while the third and fourth rows show a shear-dominant indentation made with a spherically-ended indenter. While the network in previous work\mbox{\cite{sim2real_iros}} showed sensible predictions for some indenters different from the one used for training, the figure shows how, in contrast, the network does not generalize well to light pressure conditions with the cylindrical indenter. In particular, the network predicted the typical force profile for a spherically-ended indenter. In addition, the shear-dominant indentation was also mispredicted, with the $y$ component of the force distribution predicted as symmetrical, which is the typical force profile in a vertical indentation.

\bibliographystyle{ieeetr}
\bibliography{references.bib}

\end{multicols}